\documentclass[10pt,twocolumn,letterpaper]{article}

\usepackage[pagenumbers]{cvpr} 

\usepackage{multirow, multicol}
\usepackage{threeparttable}
\usepackage{booktabs}
\usepackage{amssymb}
\usepackage{bbding}
\usepackage{pifont}
\usepackage{wasysym}
\usepackage{utfsym}
\usepackage{algorithm}
\usepackage{algpseudocode}
\usepackage[hang]{footmisc}
\usepackage{amsthm}
\usepackage[english]{babel}

\usepackage[misc]{ifsym}
\definecolor{cvprblue}{rgb}{0.21,0.49,0.74}
\usepackage[pagebackref,breaklinks,colorlinks,allcolors=cvprblue]{hyperref}

\title{Noise-Consistent Siamese-Diffusion for Medical Image \\ Synthesis and Segmentation}

\author{
    \textbf{Kunpeng Qiu}$^{1,2}$\footnote[2]{},
    \textbf{Zhiqiang Gao}$^{3}$\footnote[1]{},
    \textbf{Zhiying Zhou}$^{1,2}$\footnote[2]{},  
    \textbf{Mingjie Sun}$^{4}$\footnote[1]{},
    \textbf{Yongxin Guo}$^{1,2,5}$\thanks{Corresponding Authors. $^{\dagger}$  \text{Equal Contribution.}} \\
    {\small $^1$National University of Singapore, $^2$National University of Singapore Suzhou Research Institute, $^3$Wenzhou-Kean University} \\  
    {\small $^4$Soochow University, $^5$City University of Hong Kong} \\
    {\tt\small kunpeng\_qiu@u.nus.edu, zgao@kean.edu, elezzy@nus.edu.sg,} \\ 
    {\tt\small mjsun@suda.edu.cn, yongxin.guo@cityu.edu.hk}
}

\begin{document}
\maketitle
\begin{abstract}\label{sec:abstract}
Deep learning has revolutionized medical image segmentation, yet its full potential remains constrained by the paucity of annotated datasets. While diffusion models have emerged as a promising approach for generating synthetic image-mask pairs to augment these datasets, they paradoxically suffer from the same data scarcity challenges they aim to mitigate. Traditional mask-only models frequently yield low-fidelity images due to their inability to adequately capture morphological intricacies, which can critically compromise the robustness and reliability of segmentation models. To alleviate this limitation, we introduce Siamese-Diffusion, a novel dual-component model comprising Mask-Diffusion and Image-Diffusion. During training, a Noise Consistency Loss is introduced between these components to enhance the morphological fidelity of Mask-Diffusion in the parameter space. During sampling, only Mask-Diffusion is used, ensuring diversity and scalability. Comprehensive experiments demonstrate the superiority of our method. Siamese-Diffusion boosts SANet’s mDice and mIoU by 3.6\% and 4.4\% on the Polyps, while UNet improves by 1.52\% and 1.64\% on the ISIC2018. Code is available at \href{https://github.com/Qiukunpeng/Siamese-Diffusion}{https://github.com/Qiukunpeng/Siamese-Diffusion}.
\end{abstract}    
\section{Introduction}
\label{sec:intro}

\begin{figure}[!t]
    \centering
    \centerline{\includegraphics[width=\columnwidth]{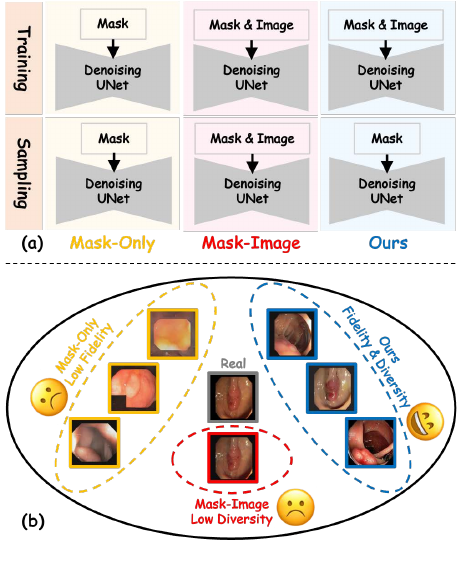}}
    \caption{(a) Workflow comparisons between our method and existing ones during the training and sampling phases. (b) Differences in synthesized images across methods. Mask-only methods (yellow) lack morphological characteristics (\eg, surface texture), resulting in low fidelity. Mask-image methods (red) produce high-fidelity images; however, their reliance on extra image prior control results in low diversity and scalability. Our method (blue) enhances morphological fidelity while preserving diversity.}
    \label{fig1}
\end{figure}

Recent advances in deep learning have led to unprecedented success in medical image analysis \cite{SANet,CTNet,ISIC2018-PAMI,Learn}. However, the full potential of segmentation remains untapped due to limited annotated datasets \cite{DiffuMask}. Unfortunately, medical data annotation is often costly, and acquiring images can be difficult or impossible due to privacy and copyright concerns. To overcome these challenges, diffusion models \cite{Score-based,DDPM,SD} have emerged as effective tools to generate image mask pairs, widely used to expand data sets for medical image segmentation tasks \cite{ArSDM,Dataset-Diffusion,DiffuseExpand}.

Typically, generating medical images involves using masks to indicate the lesion areas as prior controls injected into the diffusion model \cite{ControlNet,T2I}. However, obtaining high-fidelity synthetic medical images relies on large-scale training datasets, creating a paradox: while generative models aim to alleviate data scarcity for segmentation models, they themselves face the same challenge.

As shown in \cref{fig1}(a), mask-only models primarily focus on aligning the lesion shape to the mask \cite{ControlNetPlus, ArSDM}, often getting trapped in local minima with low morphological fidelity. Consequently, the medical images synthesized, highlighted by yellow boxes in \cref{fig1}(b), frequently neglect the indispensable morphological characteristics (\eg, surface texture) \cite{Dont,Self-guided}. This neglect obscures the discriminative features of the lesions learned by the segmentation models, rendering them unexplainable and elusive, which significantly undermines the reliability of the enhanced segmentation models.

Inspired by \cite{Classifier,Classifier-free}, incorporating additional prior guidance through the combination of images and masks significantly enhances the morphological fidelity of the synthetic images. A picture is worth a thousand words: these images contain intricate morphological information that even experienced clinicians often struggle to describe accurately. While the joint prior control substantially enhances the fidelity of synthesized morphological characteristics (marked with the red box in \cref{fig1}(b)), its close resemblance to real data (marked with the gray box in \cref{fig1}(b)) leads to a disastrous reduction in diversity. Furthermore, the scalability of the synthesized images is constrained by the scarcity of sample pairs, ultimately failing to genuinely address the data scarcity issue for downstream segmentation models.

To overcome the limitations of the aforementioned paradigms and capitalize on their strengths, we propose the resource-efficient \textsl{Siamese-Diffusion} model, which trains the same diffusion model under varying prior controls. When guided by a mask alone, this process is termed \textsl{Mask-Diffusion}, and when guided by both the image and its corresponding mask, it is called \textsl{Image-Diffusion}. To alleviate the inherent issues of \textsl{Image-Diffusion}, we introduce a \textsl{Noise Consistency Loss}, enabling the noise predicted by \textsl{Image-Diffusion} to act as an additional anchor, steering the convergence trajectory of \textsl{Mask-Diffusion} toward a local minimum with higher morphological fidelity in the parameter space. Ultimately, only \textsl{Mask-Diffusion} is employed during the sampling phase, where the diversity limitation is mitigated by the flexibility of using arbitrary masks. Our contributions are summarized as follows:
\begin{itemize}
\item[$\bullet$]
We identify that both current generative models and medical image segmentation models encounter the same data scarcity issue, leading to low morphological fidelity. We argue that only synthesized medical images containing essential morphological characteristics can ensure the reliability of enhanced segmentation models.

\item[$\bullet$]
We propose the \textsl{Siamese-Diffusion} model, which leverages \textsl{Noise Consistency Loss} and \textsl{Image-Diffusion} to guide \textsl{Mask-Diffusion} towards a local minimum with high morphological fidelity. During sampling, only \textsl{Mask-Diffusion} is used to synthesize medical images with realistic morphological characteristics.

\item[$\bullet$]
Extensive experiments demonstrate that our method outperforms existing approaches in both image quality and segmentation performance. SANet improves mDice and mIoU by 3.6\% and 4.4\% on the Polyps dataset, and UNet achieves gains of 1.52\% and 1.64\% on the ISIC2018 dataset, highlighting the superiority of our method.
\end{itemize}

\section{Related Work}\label{sec:rel}

\subsection{Controllable Diffusion Models}
In recent years, Diffusion Probabilistic Models \cite{Nonequilibrium,Score-based,DDPM} have become dominant in high-quality image synthesis. Latent Diffusion Models \cite{SD}, which later evolved into Stable Diffusion, significantly enhance both the speed and quality of image generation by mapping the diffusion process into a latent feature space \cite{VQGAN}. Furthermore, controllable conditions, such as classifier guidance \cite{Classifier,Classifier-Generating}, text guidance \cite{Classifier-free}, and structure-image guidance \cite{ControlNet,T2I}, have introduced novel approaches for advancing foundational model research. These techniques have found widespread applications in data-scarce fields, such as medical diagnosis \cite{ArSDM,Dont} and industrial anomaly detection \cite{Vos,Dream,Anomalydiffusion}.

\subsection{ReferenceNet}
Prompts often struggle to precisely control fine details and low-level semantics in generated images \cite{SD}. To address this challenge, several models \cite{Prompt-free,Reference-Animate,Reference-Magicanimate,Reference-Improving} replace complex prompts with exemplar images. These models leverage external hypernetworks \cite{Hypernetworks} to extract rich semantic information, enabling finer control over image generation. However, they prioritize consistency over diversity, limiting their applicability in tasks like medical image segmentation, which demands both morphological diversity and fidelity. Additionally, while some methods synthesize abnormal samples from normal images \cite{Controlpolypnet,Tumor}, they still raise ethical concerns. Meanwhile, adopting the IP-Adapter paradigm \cite{IP-Adapter} requires a correlation between the image prior control and noisy images (\eg, color or elements) during training, which is impractical for scarce RGB datasets. In contrast, the mask-only method offers a more accessible and streamlined framework, making it a compelling alternative for broader adoption \cite{ArSDM,DiffuseExpand,PolypDDPM}.

\begin{figure*}[!t]
    \centering
    \centerline{\includegraphics[width=\textwidth]{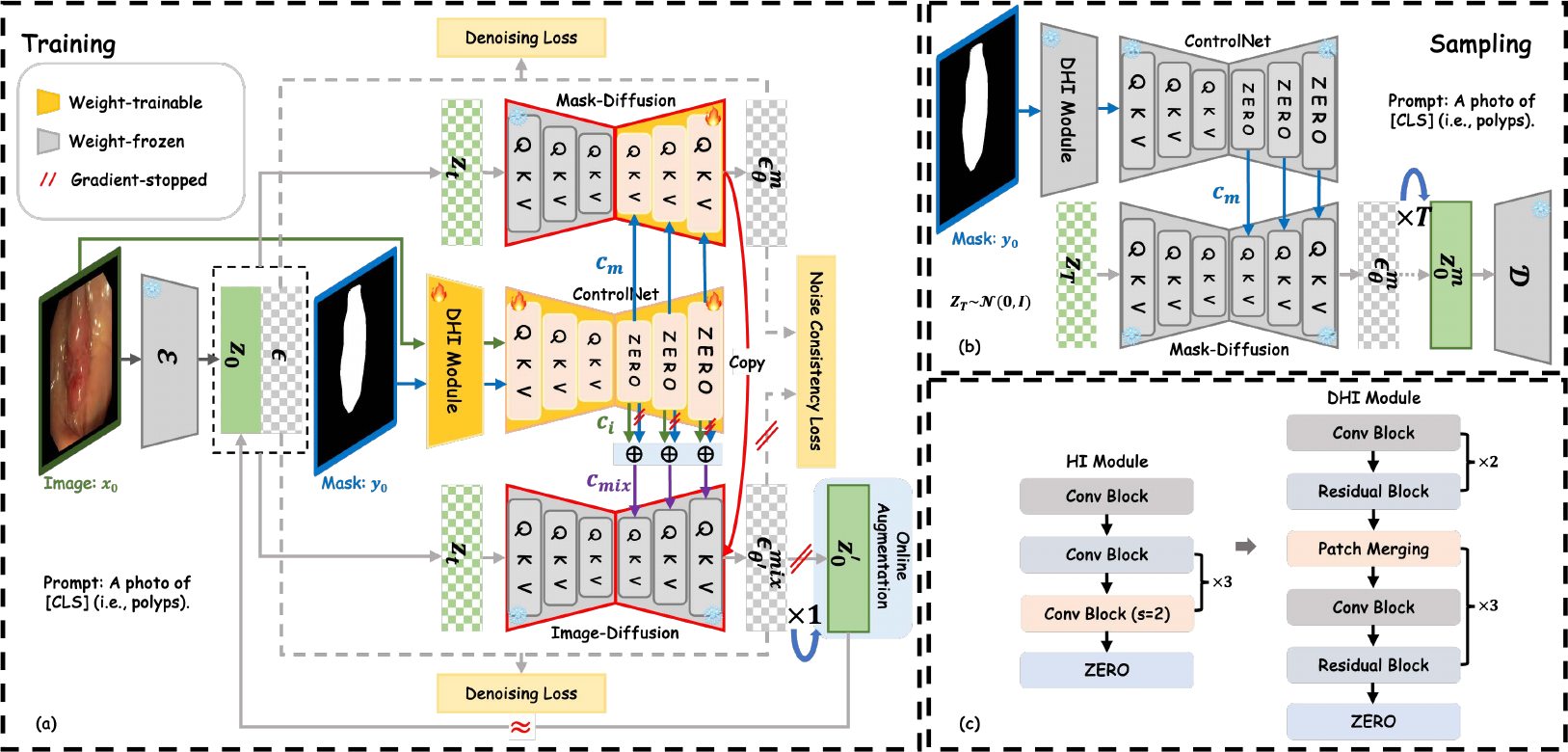}}
    \caption{(a) Illustration of our method during the training phase. The noisy image input $z_t$ in the latent space is processed through the same diffusion model (\emph{i.e.}, ``Copy") under the mask and image-mask conditions, generating the noise predictions $\epsilon_{\theta}^{m}$ and $\epsilon_{\theta^{\prime}}^{mix}$, respectively. These two processes are referred to as Mask-Diffusion and Image-Diffusion. The entire framework is optimized using the denoising losses from both cases and the proposed \textsl{Noise Consistency Loss} between $\epsilon_{\theta}^{m}$ and $\epsilon_{\theta^{\prime}}^{mix}$. The \textsl{Online-Augmentation} module employs single-step sampling to obtain the denoised $z_0^{\prime}$, which is then recombined with the mask $y_0$ to train the Mask-Diffusion. (b) During the sampling phase, only Mask-Diffusion is utilized to generate high-fidelity and diverse synthetic images. (c) Replacement of the Hint Input (HI) module with the proposed \textsl{Dense Hint Input} (DHI) module enhances the extraction of prior guidance from the image.}
    \label{fig2}
\end{figure*}

\section{Method}\label{sec:method}
\subsection{Preliminary}
Diffusion models \cite{DDPM, DDIM} comprise a diffusion process and a denoising process. The diffusion process $q(x_{1:T} | x_0)$  constitutes a fixed \textbf{T}-timestep Markov chain devoid of trainable parameters. The original input $x_0$ is iteratively perturbed by Gaussian noise according to a variance schedule $\left\{\beta_t\right\}_{t=1:T}$:
\begin{equation}\label{eq1}
    x_t = \sqrt{\Bar{\alpha}_t}x_0 + \sqrt{1 - \Bar{\alpha}_t}\epsilon, \quad \epsilon\sim\mathcal{N}(\textbf{0}, \textbf{I}),
\end{equation}
where $\epsilon$ denotes random noise sampled from a Gaussian distribution, $\alpha_t=1-\beta_t$, and $\Bar{\alpha}_t=\prod_{t=1}^{T}\alpha_t$.

Conversely, the denoising process $p_\theta(x_{t-1} | x_t)$ is trained to reverse the parameterized Markov noise process, transforming the noised Gaussian $p(x_T)$ at timestep $\textbf{T}$ into a reconstruction of the original data $x_0$, as follows:
\begin{equation}\label{eq2}
    p_{\theta}(x_{t-1} | x_t) = \mathcal{N} (x_{t-1}; \mu_{\theta}(x_t, t),\Sigma_\theta(x_t, t)),
\end{equation}
where $\mu_\theta(x_t, t)$ represents the network-predicted mean of the denoised signal at timestep $t$, $\Sigma_\theta(x_t, t)$ denotes the variance of the Gaussian distribution, which can be fixed to a constant. For simplicity, Ho et al. \cite{DDPM} parameterized this model as the function $\epsilon_\theta(x_t, t)$, predicting the noise component of a noisy sample $x_t$ at any timestep $t$. The training objective is to minimize the mean squared error loss between the actual noise $\epsilon$ and the predicted noise $\epsilon_\theta(x_t, t)$:
\begin{equation}\label{eq3}
    \mathcal{L} = \mathbb{E}_{x_t, t, \epsilon\sim\mathcal{N}(\textbf{0}, \textbf{I})} \left[ ||\epsilon_\theta(x_t, t) - \epsilon||^2 \right].
\end{equation}

Subsequently, Stable Diffusion \cite{SD} employs a pre-trained VQ-VAE \cite{VQVAE} to encode images into the latent space, conducting training on the latent representation $z_0$. In the context of controlled generation, given a text prompt $c_t$ and task-specific conditions $c_f$, the diffusion training loss at time step  $t$ can be reformulated as:
\begin{equation}\label{eq4}
    \mathcal{L} = \mathbb{E}_{z_t, t, c_t, c_f, \epsilon\sim\mathcal{N}(\textbf{0},\textbf{I})} \left[ ||\epsilon_\theta(z_t, t, c_t, c_f) - \epsilon||^2 \right].
\end{equation}

\subsection{Architecture of Siamese-Diffusion}\label{sec:architecture}

ControlNet \cite{ControlNet} and the pre-trained Stable Diffusion \cite{SD} serve as the foundational framework for our method. Following the approach outlined in \cite{ControlNet}, both the VQ-VAE module and the denoising U-Net encoder module within the Stable Diffusion framework are kept frozen. 

During the training phase, as illustrated in \cref{fig2}(a), the image $x_0$ is compressed into the latent space $z_0$ using the VQ-VAE encoder $\mathcal{E}$. The proposed \textsl{Dense Hint Input} (DHI) and ControlNet are combined in series to form a feature extraction network. This network encodes the input image $x_0$ and its corresponding mask $y_0$ into $c_i$ and $c_m$. Different prior controls, $c_m$ and $c_{mix}$, are then injected into the denoising U-Net decoder of the same diffusion model (\emph{i.e.}, ``Copy" in \cref{fig2}(a)) to predict the noise $\epsilon_{\theta}^{m}$ and $\epsilon_{\theta^{\prime}}^{mix}$, respectively. Here, $c_{mix}$ is the mixed features of $c_i$ and $c_m$:
\begin{equation}\label{eq5}
    c_{mix} = w_i \cdot c_i + w_m \cdot \textsl{sg}[c_m],
\end{equation}
where $w_i = \frac{k}{N_{iter}}$ and $w_m$ denote the weights for the image and mask prior controls. $N_{iter}$ is the total number of training iterations, and $k$ is the current iteration. For brevity, processes utilizing mask-only (\emph{i.e.}, $c_m$) and mask-image (\emph{i.e.}, $c_{mix}$) are referred to as \textsl{Mask-Diffusion} and \textsl{Image-Diffusion}. In \cref{eq5}, $c_m$ is the same as that used in \textsl{Mask-Diffusion} with truncated gradients. This Siamese architecture distinguishes it from knowledge distillation models \cite{Distilling, DistillationClassifier}, which are resource-intensive due to the need for training both a teacher and a student network. Ultimately, the loss employed to train \textsl{Siamese-Diffusion} is defined as:
\begin{equation}\label{eq6}
    \mathcal{L} = \mathcal{L}_m + \mathcal{L}_i + \mathcal{L}_c + \mathcal{L}_{m^{\prime}},
\end{equation}
where $\mathcal{L}_{c}$ is the proposed \textsl{Noise Consistency Loss}, and $\mathcal{L}_{m^{\prime}}$ arises from \textsl{Online-Augmentation}, both of which will be elaborated upon subsequently. $\mathcal{L}_{m}$ and $\mathcal{L}_{i}$ are the Denoising Losses for \textsl{Mask-Diffusion} and \textsl{Image-Diffusion}:
\begin{equation}\label{eq7}
    \mathcal{L}_{m} = \mathbb{E} \left[ || \epsilon_{\theta}^{m} - \epsilon||^2 \right], \\
    \epsilon_{\theta}^{m} = \epsilon_\theta(z_t, t, c_t, c_m),
\end{equation}
\begin{equation}\label{eq8}
    \mathcal{L}_{i} = \mathbb{E} \left[ || \epsilon_{\theta^{\prime}}^{mix} - \epsilon||^2 \right], \\
    \epsilon_{\theta^{\prime}}^{mix} = \epsilon_{\theta^{\prime}}(z_t, t, c_t, c_{mix}), \\
\end{equation}
where $\theta$ is the \textsl{Mask-Diffusion} parameter, and $\theta^{\prime}$ is a DeepCopy of $\theta$. \cref{eq7} and \cref{eq8} share the same $\epsilon \sim \mathcal{N}(\mathbf{0}, \mathbf{I})$.

During the sampling phase, as illustrated in \cref{fig2}(b), only \textsl{Mask-Diffusion} is utilized with arbitrary mask prior control to synthesize medical images for segmentation.

\begin{figure}[!t]
    \centering
    \centerline{\includegraphics[width=0.8\columnwidth]{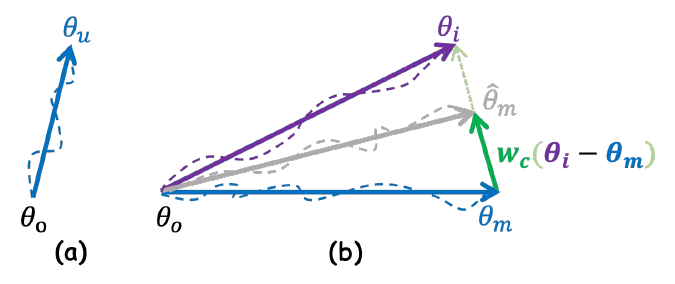}}
    \caption{(a) Parameter update direction. (b) \textsl{Mask-Diffusion} parameter update direction, scaled by the \textsl{Noise Consistency Loss}.}
    \label{fig3}
\end{figure}

\subsection{Noise Consistency Loss}
As mentioned above, in data-scarce scenarios, \textsl{Mask-Diffusion} tends to get trapped in low-fidelity local minima, yielding synthesized medical images that lack indispensable morphological characteristics, such as surface texture. In contrast, \textsl{Image-Diffusion}, aided by the additional image prior control $x_0$, more effortlessly converges to high-fidelity local optima. However, during the sampling phase, the strong prior control from the image $x_0$, which leads to synthesized images closely resembling real images as shown in \cref{fig1}(b), restricts the diversity of morphological characteristics and limits the scalability due to the scarcity of paired samples. As illustrated in \cref{tab2}, both cases lead to catastrophic degradation of segmentation performance.

To help \textsl{Mask-Diffusion} escape from low-fidelity local minima and move toward higher-fidelity regions in the parameter space, we introduce the \textsl{Noise Consistency Loss}:
\begin{equation}\label{eq9}
    \mathcal{L}_{c} =  w_c \cdot \mathbb{E} \left[ ||\epsilon_{\theta}^m - \textsl{sg}[\epsilon_{\theta^{\prime}}^{mix}]||^2 \right],
\end{equation}
where $\epsilon_{\theta^{\prime}}^{mix}$, the predicted noise from \textsl{Image-Diffusion}, is more accurate than $\epsilon_{\theta}^m$ predicted by \textsl{Mask-Diffusion}, owing to the additional image prior control. The stop-gradient operation \cite{SimSiam} $\textsl{sg}$ enables the more accurate $\epsilon_{\theta^{\prime}}^{mix}$ to serve as an anchor, guiding the convergence trajectory of \textsl{Mask-Diffusion} towards higher-fidelity local minima in the parameter space. The ``Copy" operation in \cref{fig2}(a) and the identical $c_m$ in \cref{eq5}, jointly ensure that the strong prior control provided by the additional image is successfully propagated to \textsl{Mask-Diffusion}. Meanwhile, $w_i = \frac{k}{N_{iter}}$ gradually increases, with Denoising Loss dominating the early stages to ensure stable convergence. $w_c$ controls the steering strength, as discussed in \cref{5.3.2}.

\begin{figure}[!t]
    \centering
    \centerline{\includegraphics[width=\columnwidth]{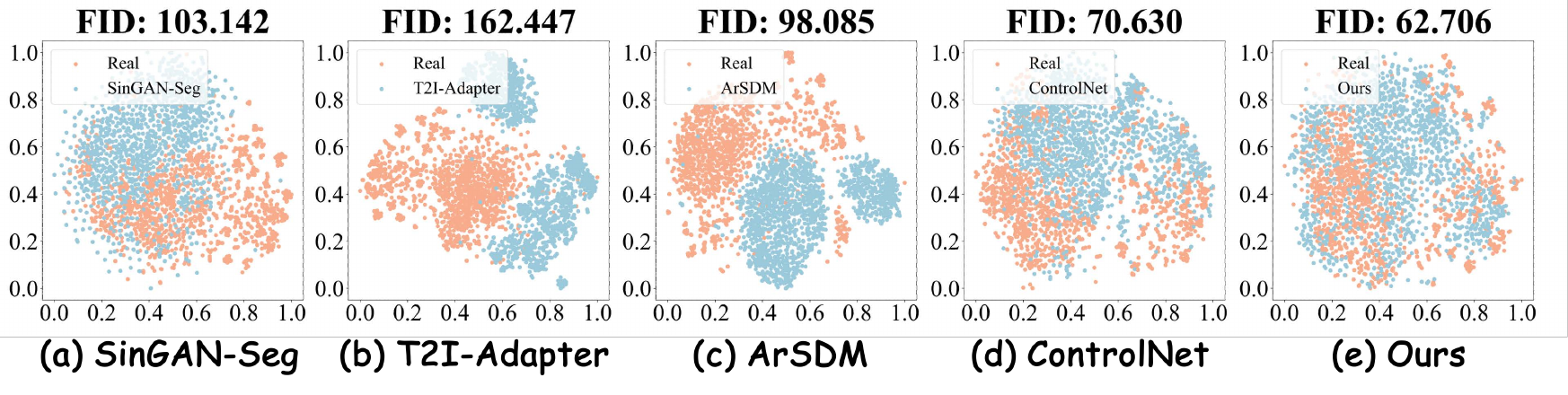}}
    \caption{t-SNE visualization of data distribution. (a)–(e) illustrate the distribution differences between real polyp images and those synthesized by each respective mask-only method. The distribution of polyp images generated by our method nearly overlaps with the real data, underscoring its exceptional ability to produce highly realistic polyp images.}
    \label{fig4}
\end{figure}

\begin{figure*}[!t]
    \centering
    \centerline{\includegraphics[width=\textwidth]{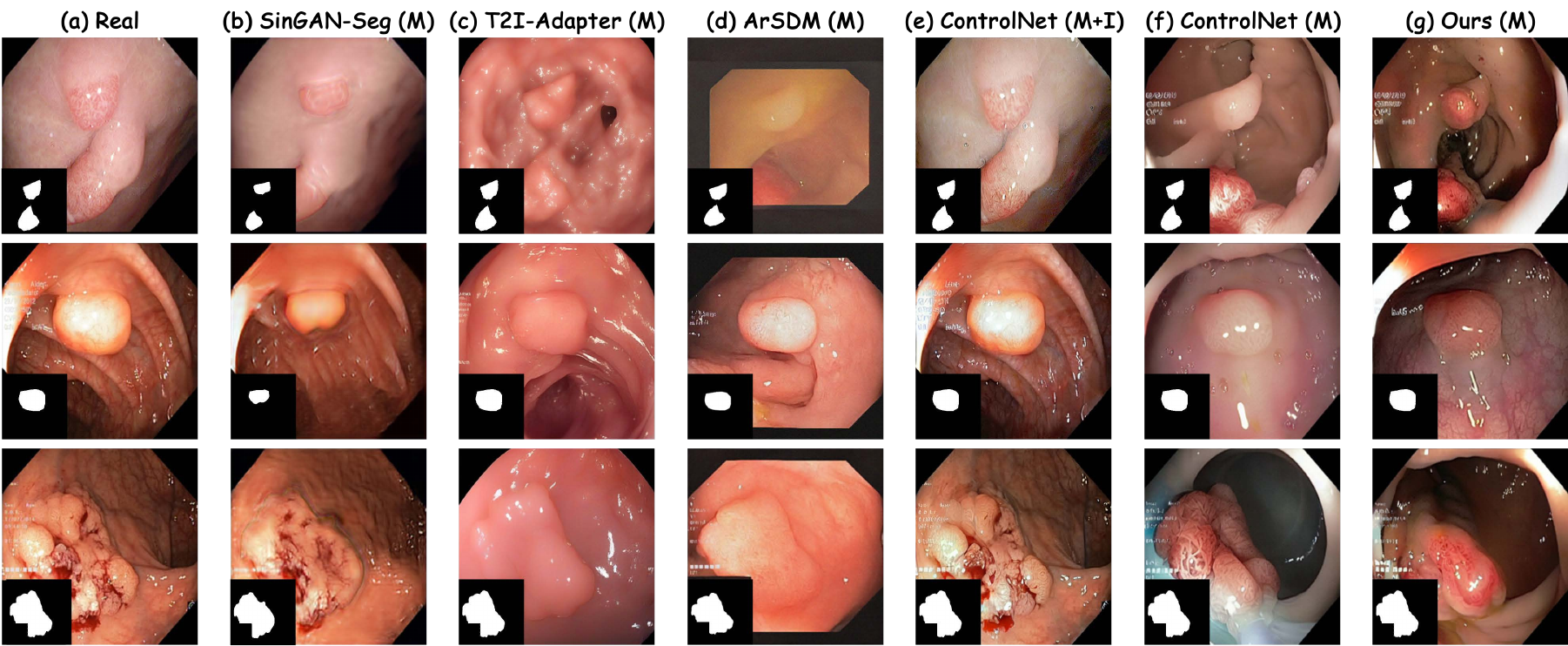}}
    \caption{(a) Examples of real polyp images. (b)–(g) Examples of synthetic polyp images generated by each respective method. ``M" denotes that mask-only prior control, while ``M+I" denotes mask-image joint prior control. The synthetic polyp images generated by our method achieve competitive morphological fidelity while also exhibiting morphological diversity (Zoom in for better visualization).}
    \label{fig5}
\end{figure*}

\cref{fig3} visually illustrate the above operation in parameter space. Using arithmetic operations \cite{Editing, Perturbation}, the update direction of the parameters can be depicted by the difference between the original model parameters $\theta_o$, and the updated model parameters $\theta_u$. Therefore, the final update direction for the \textsl{Mask-Diffusion} parameters can be approximated as:
\begin{equation}\label{eq10}
\hat{{\theta}}_m = \theta_m + w_c \cdot (\theta_i - \theta_m),
\end{equation}
where $\theta_m$ and $\theta_i$ are the parameters of \textsl{Mask-Diffusion} and \textsl{Image-Diffusion}, and $\hat{{\theta}}_m$ is the refined \textsl{Mask-Diffusion} parameter. Throughout this process, collaborative updates to the feature extraction network and the diffusion model maintain the alignment of mask features $c_m$ with noisy image features $z_t$ in the latent space, ensuring the fused latent features reside within a suitable manifold. Consequently, \textsl{Mask-Diffusion} can operate independently during sampling. As shown in \cref{fig1}(b), the morphological characteristics of the synthesized images are enhanced by the refined \textsl{Mask-Diffusion}, making these images competitive with those synthesized by \textsl{Image-Diffusion}. Segmentation models enhanced with these images ensure interpretability and reliability. Unlike the classifier-free guidance (CFG) \cite{Classifier-free}, which improves fidelity by adjusting guidance strength in the sample space, our proposed \textsl{Siamese-Diffusion} enhances morphological fidelity within the parameter space. This distinction eliminates the limitations imposed by the reliance on paired controls during sampling.

\subsection{Dense Hint Input Module}
The prior controls used in \cite{ControlNet} are typically low-density semantic images, such as segmentation masks, depth maps, and sketches, from which features can be effortlessly extracted using a sparse Hint Input (HI) module. However, when handling high-density semantic images, such as medical images, the original sparse HI module is insufficient for capturing nuanced details, including texture and color. Thus, a \textsl{Dense Hint Input} (DHI) module is introduced. As illustrated in \cref{fig2}(c), the DHI module incorporates denser residual blocks with channel sizes of 16, 32, 64, 128 and 256, along with a more advanced patch merging module \cite{Swin}. This design enable the DHI module effectively accommodates both image and mask prior controls \cite{Prompt-free}.

\subsection{Online-Augmentation}
Leveraging the advantages of the \textsl{Siamese-Diffusion} training paradigm, \textsl{Online-Augmentation} is introduced to expand the \textsl{Mask-Diffusion} training set, as shown in \cref{fig2}(a). The additional image prior control $x_0$ improves the accuracy of noise predicted by \textsl{Image-Diffusion}, enabling single-step sampling \cite{ControlNetPlus, DDIM} to generate $z_0^{\prime}$ online. In the first step, $z_0^{\prime}$ is obtained through the reverse process of \cref{eq1}, with $x_t$ replaced by $z_t$, as follows:

\begin{equation}\label{eq11}
z_0 \approx z_0^{\prime} = \frac{z_t - \sqrt{1 -\Bar{\alpha}_t} \textsl{sg}[\epsilon_{\theta^{\prime}}^{mix}]}{\sqrt{\Bar{\alpha}_t}},
\end{equation}
where $\textsl{sg}[\epsilon_{\theta^{\prime}}^{mix}]$ is the noise predicted by \textsl{Image-Diffusion} with truncated gradients. In the second step, $z_0^{\prime}$ approximates $z_0$, recombining with $c_m$ to train \textsl{Mask-Diffusion}:

\begin{table}[!t]\small
    \centering
    \caption{Comparison of synthetic polyp image quality generated by each respective mask-only method, evaluated using FID \cite{FID}, KID \cite{KID}, CLIP-I \cite{CLIP-I}, LPIPS \cite{LPIPS}, CMMD \cite{CMMD}, and MOS metrics.}
    \label{tab1}
    \setlength{\tabcolsep}{0.1mm}
    \scalebox{0.7}
    {\begin{tabular}{l|c c c c c c}
        \toprule[2pt]
        Methods & FID ($\downarrow$) & KID ($\downarrow$) & CLIP-I ($\uparrow$) & LPIPS ($\downarrow$) & CMMD ($\downarrow$) & MOS (Confidence) ($\uparrow$)\\
        \midrule[0.75pt]
        SinGAN-Seg \cite{Singan} & 103.142 & 0.0898 & 0.851 & 0.639 & 0.929 & 0.233 (8.25) \\
        T2I-Adapter \cite{T2I} & 162.447 & 0.1517 & 0.875 & 0.672 & 0.797 & -\\
        ArSDM \cite{ArSDM} & 98.085 & 0.0927 & 0.845  & 0.727 & 0.811 & -\\
        ControlNet \cite{ControlNet} & 70.630 & 0.0509 & 0.887 & 0.612 & 0.543 & 0.487 (5.94)\\
        \textbf{Ours} & \textbf{62.706} & \textbf{0.0395} & \textbf{0.892} & \textbf{0.586} & \textbf{0.515} & \textbf{0.587 (6.04)}\\
        \bottomrule[2pt]
    \end{tabular}}
\end{table}

\begin{equation}\label{eq12}
\mathcal{L}_{m^{\prime}} = w_a \cdot \mathbb{E} \left[ ||\epsilon_\theta^{m^{\prime}} - \epsilon||^2 \right], \epsilon_\theta^{m^{\prime}}=\epsilon_\theta(z_t^{\prime}, t, c_t, c_m),
\end{equation}
where $\epsilon$ is as in \cref{eq7}, and $w_a$ controls alignment between $z_0^{\prime}$ and $c_m$, following \cite{ControlNetPlus}:
\begin{equation}\label{eq13}
    w_a = \begin{cases} 
    1, & \text{if } k > K_{\tau} \text{ and } t < T_{\tau}, \\
    0, & \text{otherwise},
    \end{cases}
\end{equation}
where $k$ is the current iteration, $K_{\tau}$ is the iteration threshold, $t$ is the timestep, and $T_{\tau}$ is the timestep threshold.

\begin{table}[t]\tiny
    \centering
    \begin{threeparttable}
    \caption{Comparison of SANet \cite{SANet}, Polyp-PVT \cite{Polyp-PVT}, and CTNet \cite{CTNet} performance, evaluated using mDice (\%) and mIoU (\%) metrics. ``+" denotes that the training dataset consists of both real polyp images and synthetic polyp images generated by each respective method. ``M" denotes that mask-only prior control, while ``M+I" denotes mask-image joint prior control.}\label{tab2}
    \setlength{\tabcolsep}{0.35mm}
    \begin{tabular}{l|c c c c c c c c c c|c c}
        \toprule[2pt]
        \multirow{2.5}{*}{Methods} & \multicolumn{2}{c}{EndoScene \cite{EndoScene}} & \multicolumn{2}{c}{ClinicDB \cite{CVC-ClinicDB}} & \multicolumn{2}{c}{Kvasir \cite{Kvasir}} & \multicolumn{2}{c}{ColonDB \cite{CVC-ColonDB}} & \multicolumn{2}{c|}{ETIS \cite{ETIS}} & \multicolumn{2}{c}{Overall}\\
        \cmidrule{2-13}
        ~ & mDice & mIoU & mDice & mIoU & mDice & mIoU & mDice & mIoU & mDice & mIoU & mDice & mIoU \\
        \midrule[0.75pt]
        SANet \cite{SANet} & 88.8 & 81.5 & 91.6 & 85.9 & 90.4 & 84.7 & 75.3 & 67.0 & 75.0 & 65.4 & 79.4 & 71.4 \\
        +Copy-Paste & 89.7 & 83.0 & 90.2 & 85.1 & 90.3 & 84.8 & 77.7 & 70.0 & 77.4 & 68.8 & 81.1 & 73.7 \\
        +SinGAN-Seg \cite{Singan} (M) & 88.3 & 81.6 & 90.9 & 85.3 & 91.0 & \textbf{85.8} & 77.3 & 69.4 & 73.7 & 65.4 & 80.0 & 72.6 \\
        +T2I-Adapter \cite{T2I} (M) & 86.5 & 78.9 & 90.6 & 85.0 & 89.9 & 84.1 & 77.7 & 69.4 & 78.6 & 69.7 & 81.1 & 73.2 \\
        +ArSDM \cite{ArSDM} (M) & 90.2 & 83.2 & 91.4 & 86.1 & 91.1 & 85.6 & 77.7 & 70.0 & 78.0 & 69.5 & 81.5 & 74.1 \\
        +ControlNet \cite{ControlNet} (M) & 90.2 & 83.7 & 91.6 & 85.4 & 90.6 & 85.0 & 77.0 & 69.0 & 75.8 & 66.5 & 80.5 & 72.8 \\
        +ControlNet \cite{ControlNet} (M+I) & 89.9 & 83.3 & 91.4 & 86.2 & \textbf{91.2} & 85.5 & 78.2 & 70.5 & 75.4 & 66.6 & 81.0 & 73.6 \\
        +\textbf{Ours} (M) & \textbf{90.4} & \textbf{83.9} & \textbf{93.0} & \textbf{88.1} & \textbf{91.2} & 85.5 & \textbf{79.1} & \textbf{71.3} & \textbf{81.1} & \textbf{73.4} & \textbf{83.0} & \textbf{75.8} \\
        \midrule[0.75pt]
        Polyp-PVT \cite{Polyp-PVT} & 90.0 & 83.3 & 93.7 & 88.9 & 91.7 & 86.4 & 80.8 & 72.7 & 78.7 & 70.6 & 83.3 & 76.0 \\
        +Copy-Paste & 88.0 & 80.9 & 93.4 & 88.7 & 91.7 & 87.1 & 79.8 & 71.8 & 79.2 & 71.3 & 82.8 & 75.6 \\
        +SinGAN-Seg \cite{Singan} (M) & 87.0 & 79.7 & 91.7 & 87.0 & \textbf{92.8} & \textbf{88.1} & 76.9 & 69.0 & 74.2 & 66.7 & 80.1 & 73.0 \\
        +T2I-Adapter \cite{T2I} (M) & 88.7 & 81.6 & 93.1 & 88.1 & 91.3 & 86.0 & 80.6 & 72.4 & 79.0 & 71.0 & 83.1 & 75.7 \\
        +ArSDM \cite{ArSDM} (M) & 88.2 & 81.2 & 92.2 & 87.5 & 91.5 & 86.3 & \textbf{81.7} & 73.8 & 80.6 & 72.9 & 84.0 & 76.7 \\
        +ControlNet \cite{ControlNet} (M) & 88.0 & 81.1 & 93.7 & 89.0 & 91.5 & 86.3 & 80.8 & 72.5 & 76.1 & 68.1 & 82.5 & 75.1 \\
        +ControlNet \cite{ControlNet} (M+I) & 90.3 & 83.5 & 92.2 & 87.7 & 91.1 & 86.2 & 81.6 & 73.7 & 77.2 & 69.1 & 83.2 & 76.0 \\
        +\textbf{Ours} (M) & \textbf{90.4} & \textbf{83.8} & \textbf{93.9} & \textbf{89.3} & 91.1 & 85.9 & \textbf{81.7} & \textbf{74.0} & \textbf{81.5} & \textbf{73.4} & \textbf{84.4} & \textbf{77.3} \\
        \midrule[0.75pt]
        CTNet \cite{CTNet} & 90.8 & 84.4 & 93.6 & 88.7 & 91.7 & 86.3 & 81.3 & 73.4 & \textbf{81.0} & 73.4 & 84.2 & 77.0 \\
        +Copy-Paste & \textbf{91.6} & \textbf{85.5} & 92.3 & 87.1 & 91.5 & 86.3 & 82.5 & 74.7 & 78.1 & 70.6 & 84.0 & 76.9 \\
        +SinGAN-Seg \cite{Singan} (M) & 89.4 & 82.3 & 93.6 & 88.8 & \textbf{92.9} & 88.6 & 78.0 & 70.4 & 76.4 & 68.7 & 81.5 & 74.6 \\
        +T2I-Adapter \cite{T2I} (M) & 90.8 & 84.3 & 93.7 & 88.7 & 91.9 & 86.9 & 81.7 & 74.1 & 80.4 & 72.5 & 84.3 & 77.2 \\
        +ArSDM \cite{ArSDM} (M) & 89.8 & 82.9 & 92.9 & 87.8 & 90.5 & 85.2 & 81.9 & 74.0 & 80.9 & \textbf{73.5} & 84.2 & 77.0 \\
        +ControlNet \cite{ControlNet} (M) & 90.2 & 84.1 & 92.3 & 87.5 & 91.7 & 86.9 & 81.4 & 73.5 & 79.5 & 71.7 & 83.7 & 76.6 \\
        +ControlNet \cite{ControlNet} (M+I) & 90.7 & 84.2 & 93.2 & 88.6 & 92.0 & 87.2 & 81.1 & 73.6 & 80.9 & 73.1 & 84.1 & 77.1 \\
        +\textbf{Ours} (M) & 90.1 & 83.6 & \textbf{93.8} & \textbf{88.9} & \textbf{92.9} & \textbf{89.1} & \textbf{83.8} & \textbf{75.8} & 79.5 & 72.0 & \textbf{85.1} & \textbf{78.1} \\
        \bottomrule[2pt]
    \end{tabular}
    \end{threeparttable}
\end{table}
\section{Experimental}\label{sec:experiment}
\subsection{Dataset and Metrics}\label{sec:dataset}
We conduct experiments on three public medical datasets—Polyps \cite{Kvasir,CVC-ClinicDB}, ISIC2016 \cite{ISIC2016}, and ISIC2018 \cite{ISIC2018}—and two internal natural datasets, Stain and Faeces.

\textbf{Training Set for Generative Models:} Following \cite{ArSDM}, the Polyps dataset consists of $1,450$ samples, comprising $900$ samples from Kvasir \cite{Kvasir} and $550$ samples from CVC-ClinicDB \cite{CVC-ClinicDB}. The ISIC2016 \cite{ISIC2016} and ISIC2018 \cite{ISIC2018} datasets contain $900$ and $2,594$ samples, respectively. The Stain ($500$ samples) and Faeces ($458$ samples) datasets are partitioned into training, validation, and test sets in a 3:1:1 ratio, yielding training sets of $300$ and $275$ samples, respectively.

\textbf{Training Set for Segmentation Models:} The medical datasets leverage synthetic images generated using masks derived from their original training sets, which are then amalgamated with the original images to create new training sets. For the Stain and Faeces datasets, the original masks are augmented through various transformations (\eg, scaling) to produce $1,000$ samples for training.

\textbf{Evaluation Metrics:} We use several image quality evaluation metrics, including FID \cite{FID}, KID \cite{KID}, CLIP-I \cite{CLIP-I}, LPIPS \cite{LPIPS}, CMMD \cite{CMMD}, and the Mean Opinion Score (MOS), which is assessed by 3 experienced clinicians (see Appendix for details). We use mDice and mIoU to evaluate segmentation performance with the default settings of CNN and Transformer-based models.

\begin{table}[!t]\small
    \centering
    \caption{Comparison of UNet \cite{UNet} and SegFormer \cite{SegFormer} performance, evaluated using mDice (\%) and mIoU (\%). ``+" denotes that the training dataset consists of both real and synthetic skin lesion images generated by each respective mask-only method. ``Real" denotes that masks are derived from real data, while ``Random" indicates transformed versions of real masks (\eg, scaling).}\label{tab3}
    \setlength{\tabcolsep}{0.4mm}
    \scalebox{0.8}
    {\begin{tabular}{l|c c|c c|c c|c c}
        \toprule[2pt]
        \multirow{4}{*}{Methods} & \multicolumn{4}{c|}{ISIC2016 \cite{ISIC2016}} & \multicolumn{4}{c}{ISIC2018 \cite{ISIC2018}} \\ 
        \cmidrule{2-9}
        ~ & \multicolumn{2}{c|}{UNet \cite{UNet}} & \multicolumn{2}{c|}{SegFormer \cite{SegFormer}} & \multicolumn{2}{c|}{UNet \cite{UNet}} & \multicolumn{2}{c}{SegFormer \cite{SegFormer}} \\ 
        \cmidrule{2-9}
        ~ & mDice & mIoU & mDice & mIoU & mDice & mIoU & mDice & mIoU \\
        \midrule[0.75pt]
        Real Dataset & 89.58 & 86.62 & 93.57 & 89.06 & 82.19 & 78.45 & 91.35 & 86.10 \\
        +Copy-Paste & 89.43 & 86.48 & 93.53 & 88.85 & 81.00 & 77.16 & 91.25 & 86.09 \\
        +T2I-Adapter \cite{T2I} & 89.48 & 86.47 & 93.71 & 89.24 & 81.16 & 77.04 & 91.19 & 85.73 \\
        +ControlNet \cite{ControlNet} & 89.42 & 86.51 & 93.54 & 88.83 & 81.76 & 77.94 & 91.52 & 86.19 \\
        +\textbf{Ours} (Real) & \underline{89.91} &  \underline{87.01} &  \underline{94.14} &  \underline{89.76} &  \underline{82.81} &  \underline{79.04} &  \underline{91.80} &  \underline{86.67} \\
        +\textbf{Ours} (Random) & \textbf{90.06} & \textbf{87.25} & \textbf{94.31} & \textbf{90.01} & \textbf{83.71} & \textbf{80.09} & \textbf{91.93} & \textbf{87.12} \\
        \bottomrule[2pt]
    \end{tabular}}
\end{table}

\begin{table}[!t]\small
    \centering
    \caption{Comparison of SegFormer \cite{SegFormer} performance, evaluated on mDice (\%) and mIoU (\%). Training dataset includes 1,000 synthetic Stain/Faeces images generated by each mask-only method.}\label{tab4}
    \setlength{\tabcolsep}{2.5mm}
    \scalebox{0.9}
    {\begin{tabular}{l|c c|c c}
        \toprule[2pt]
        \multirow{4}{*}{Methods} & 
        \multicolumn{2}{c|}{Stain} & \multicolumn{2}{c}{Faeces}\\
        \cmidrule{2-5}
        ~ & \multicolumn{2}{c|}{SegFormer \cite{SegFormer}} & \multicolumn{2}{c}{SegFormer \cite{SegFormer}} \\
        \cmidrule{2-5}
        ~ & mDice & mIoU & mDice & mIoU \\
        \midrule[0.75pt]
        Real Dataset & 93.46 & 89.78 & 90.50 & 87.78 \\
        \hline
        DFMGAN \cite{DFMGAN} & 49.59 & 49.19 & 49.47 & 48.95 \\
        AnomalyDiffusion \cite{Anomalydiffusion} & 70.84 & 64.47 & 81.59 & 73.39 \\
        T2I-Adapter \cite{T2I} & 74.75 & 70.57 & 88.02 & 83.88\\
        ControlNet \cite{ControlNet} & 76.65 & 70.48 & 88.50 & 83.58 \\
        \textbf{Ours} & \textbf{87.96} & \textbf{83.37} & \textbf{94.11} & \textbf{90.26} \\
        \bottomrule[2pt]
    \end{tabular}}
\end{table}

\subsection{Implementation Details}
Pre-trained Stable Diffusion V1.5 \cite{SD} forms the foundation of \textsl{Siamese-Diffusion}. We fine-tune Stable Diffusion across all five datasets using the AdamW \cite{AdamW} optimizer, with a learning rate of $1 \times 10^{-5}$ and a weight decay of $1 \times 10^{-2}$. The maximum number of training iterations (\emph{i.e.}, $N_{iter}$) is set to $3,000$ for the Polyps \cite{Kvasir,CVC-ClinicDB}, ISIC2016 \cite{ISIC2016}, and ISIC2018 \cite{ISIC2018} datasets. It is set to $1,500$ for the Stain and Faeces datasets. In \cref{eq5}, the weight $w_m$ is set to $1.0$. In \cref{eq13}, the iteration threshold $K_{\tau} = \frac{N_{iter}}{3}$ is empirically chosen and the timestep threshold $T_{\tau} = 200$ follows \cite{ControlNetPlus}. We use an optimal value of $w_c = 1.0$, as elaborated in \cref{5.3.2}. The images of the five datasets are resized to $384\times384$, and $5\%$ probability is applied with no prompt \cite{Classifier-free, ControlNet}. We execute training on $8$ NVIDIA 4090 GPUs with a batch size of $6$, culminating in a total batch size of $48$. During sampling, we utilize DDIM \cite{DDIM} ($\eta=0$) with $50$ steps and a guidance scale of $\lambda = 9$, in accordance with the methodology in \cite{ControlNet}.

\section{Evaluations and results}
\subsection{Image Quality Comparison}\label{sec:imagequality}
We evaluate synthetic images from prior mask-only models quantitatively and qualitatively on the Polyps dataset, with results from other datasets provided in the Appendix.

\textbf{Quantitative Evaluation:} 
Our method outperforms others across six metrics, achieving a $35$-point reduction in FID compared to the previous SOTA ArSDM \cite{ArSDM}, as shown in \cref{tab1}. The t-SNE visualizations in \cref{fig4} further substantiate the realism of our synthetic images. Additionally, our method achieves a KID score of $0.0395$ and a CLIP-I rating of $0.892$, reflecting superior image quality. Meanwhile, the best LPIPS score of $0.586$ and CMMD score of $0.515$ demonstrate a closer alignment with human perception, while the top MOS score of $0.587$ supports this conclusion, highlighting enhanced morphological characteristics.

\textbf{Qualitative Evaluation:} 
\cref{fig5} presents polyp images generated by various methods. SinGAN-Seg \cite{Singan} often introduces artifacts and lacks diversity due to its ``editing-like" approach. T2I-Adapter \cite{T2I} produces a ``glossy" appearance, leading to unrealistic images. ArSDM \cite{ArSDM} fails to synthesize key morphological characteristics (\eg, surface texture), resulting in low fidelity. Our method, based on ControlNet \cite{ControlNet}, outperforms the mask-only (\emph{i.e.}, ``M") ControlNet \cite{ControlNet} shown in \cref{fig5}(f) in three critical areas: mask alignment, morphological texture, and color, demonstrating its superiority. Compared to the mask-image (\emph{i.e.}, ``M+I") ControlNet \cite{ControlNet} shown in \cref{fig5}(e), our method exhibits competitive fidelity while preserving diversity.

\begin{table}[!t]\small
    \centering
    \caption{Comparison of the impact of different components on polyp image quality and segmentation performance. Image quality is evaluated using FID \cite{FID}, KID \cite{KID}, and LPIPS \cite{LPIPS} metrics. Segmentation performance is evaluated using weighted mDice (\%) and mIoU (\%) metrics across five public test datasets. The training dataset consists of real polyp images and images synthesized by our proposed \textsl{Siamese-Diffusion} method for each setting.}\label{tab5}
    \setlength{\tabcolsep}{0.5mm}
    \scalebox{0.6}
    {\begin{tabular}{c|c c c|c c c|c c c c c c}
        \toprule[2pt]
        \multirow{2.5}{*}{Settings} & \multirow{2.5}{*}{DHI} & \multirow{2.5}{*}{Online-Aug} & \multirow{2.5}{*}{$\mathcal{L}_{s}$} & \multirow{2.5}{*}{FID ($\downarrow$)} & \multirow{2.5}{*}{KID ($\downarrow$)} & 
        \multirow{2.5}{*}{LPIPS ($\downarrow$)} & \multicolumn{2}{c}{SANet \cite{SANet}} & \multicolumn{2}{c}{Polyp-PVT \cite{Polyp-PVT}} & \multicolumn{2}{c}{CTNet \cite{CTNet}} \\
        \cmidrule{8-13}
        ~ & ~ & ~ & ~ & ~ & ~ & ~ & mDice & mIoU & mDice & mIoU & mDice & mIoU \\
        \midrule[0.75pt]
        1 & ~ & ~ & ~ & 70.630 & 0.0509 & 0.612 & 80.5 & 72.8 & 82.5 & 75.1 & 83.7 & 76.6 \\
        2 & \CheckmarkBold & ~ & ~ & 70.479 & 0.0496 & 0.610 & 80.8 & 73.3 & 83.1 & 75.9 & 83.8 & 76.8 \\
        3 & ~ & \CheckmarkBold & ~ & 65.927 & 0.0459 & 0.605 & 80.9 & 73.4 & 82.7 & 75.5 & 84.0 & 76.9 \\
        4 & ~ & ~ & \CheckmarkBold & 67.508 & 0.0476 & 0.602 & 81.8 & 74.6 & 82.8 & 75.6 & 84.3 & 77.1 \\
        5 & ~& \CheckmarkBold & \CheckmarkBold & 65.266 & 0.0458 & 0.601 & 82.0 & 74.4 & 83.7 & 76.5 & 84.4 & 77.3 \\
        6 & \CheckmarkBold & \CheckmarkBold  & ~ & \textbf{62.706} & \textbf{0.0385} & 0.598 & 81.7 & 74.2 & 83.5 & 76.4 & 84.1 & 77.1 \\
        7 & \CheckmarkBold & ~ & \CheckmarkBold & 63.059 & 0.0405 & 0.590 & 82.7 & 75.2 & 84.2 & 76.8 & 84.7 & 77.7 \\
        \textbf{Ours} & \CheckmarkBold & \CheckmarkBold & \CheckmarkBold & \textbf{62.706} & 0.0395 & \textbf{0.586}& \textbf{83.0} & \textbf{75.8} & \textbf{84.4} & \textbf{77.3} & \textbf{85.1} &\textbf{78.1} \\
        \bottomrule[2pt]
    \end{tabular}}
\end{table}

\subsection{Segmentation Performance Comparison}\label{Segmentation Comparison}
Polyp segmentation performance comparisons are provided in \cref{tab2}. We retrain models on a duplicated dataset (\emph{i.e.}, ``Copy-Paste") to establish a new baseline. Our method surpasses the previous SOTA, ArSDM \cite{ArSDM}, significantly improving mDice and mIoU for SANet \cite{SANet} by 3.6\% and 4.4\%. For more advanced architectures, such as Polyp-PVT \cite{Polyp-PVT} and CTNet \cite{CTNet}, mDice scores increase by $1.1\%$ and $0.9\%$, and mIoU increases by $1.3\%$ and $1.1\%$. Only minor declines are observed on Kvasir with Polyp-PVT and on EndoScene and ETIS with CTNet, highlighting the robustness of our method across diverse models and test sets.

Intriguingly, despite ControlNet’s \cite{ControlNet} superior morphological quality with mask-image (\emph{i.e.}, ``M+I") joint prior controls, the segmentation performance of Polyp-PVT and CTNet declines, similar to Copy-Paste. Conversely, ArSDM \cite{ArSDM}, despite lower morphological fidelity, improves segmentation performance. These observations reveal a crucial insight: image quality alone does not solely determine segmentation performance; morphological diversity is equally essential. However, we emphasize that, in medical imaging applications, morphological diversity must be paired with high morphological fidelity to ensure interpretable and reliable segmentation.

Similar phenomena can be observed in \cref{tab3}. Interestingly, both UNet \cite{UNet} and SegFormer \cite{SegFormer} perform better with skin lesion images synthesized using random masks (\emph{i.e.}, ``Random") compared to those derived from real data (\emph{i.e.}, ``Real"). UNet achieves a remarkable increase of $1.52\%$ in mDice and $1.64\%$ in mIoU on ISIC2018, underscoring the sensitivity of segmentation to morphological diversity. \cref{tab4} presents segmentation performance comparisons using $1,000$ Stain or Faeces samples synthesized with transformed masks (\emph{e.g.}, scaling), without real samples. Our method significantly outperforms AnomalyDiffusion \cite{Anomalydiffusion}, even surpassing results achieved with real faeces samples, highlighting its superior scalability.

\begin{figure}[!t]
    \centering
    \centerline{\includegraphics[width=\columnwidth]{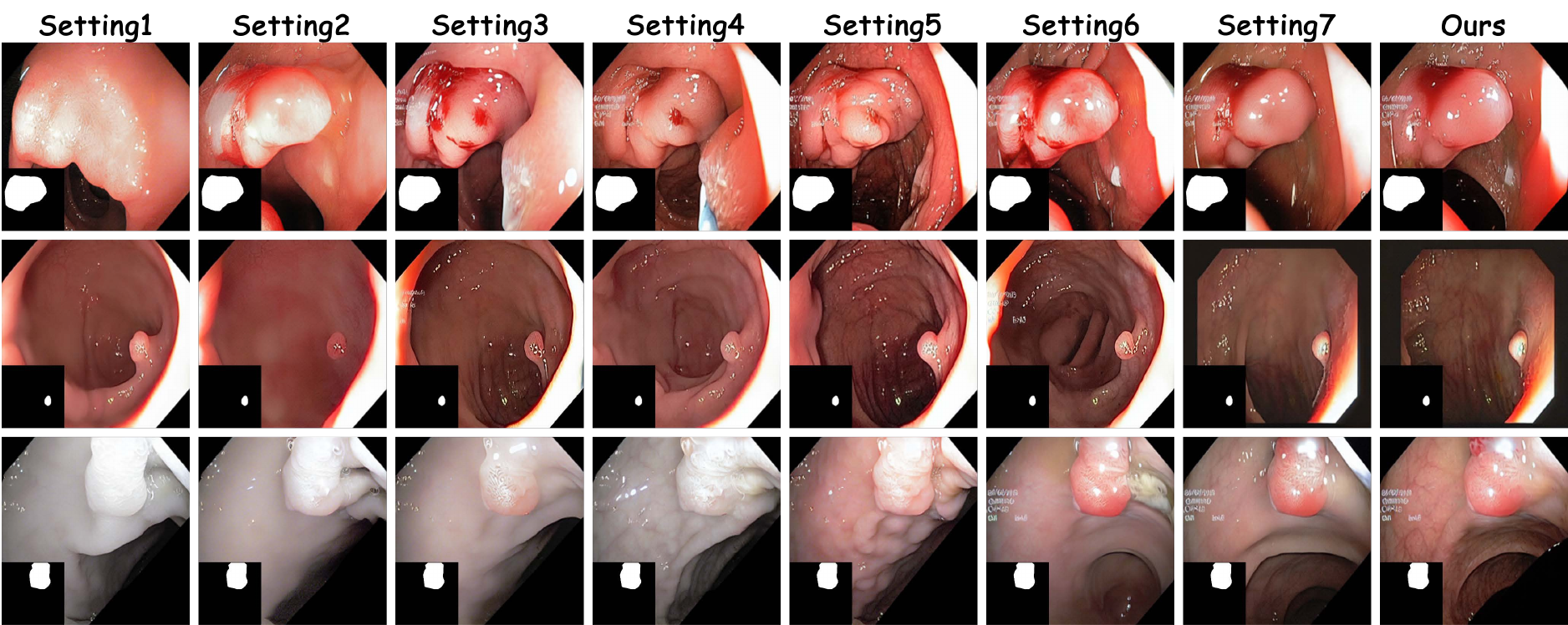}}
    \caption{Visualizing the impact of different components on the synthesis of polyp images (Zoom in for better visualization).}
    \label{fig6}
\end{figure}

\subsection{Ablation Studies}
\subsubsection{Contribution of Main Components}\label{components}
We conduct the first ablation study to evaluate the contribution of each main component. 

\textbf{Quantitative Analysis:} As illustrated in \cref{tab5}, Setting 1 corresponds to ControlNet \cite{ControlNet}. Setting 2 introduces DHI, which improves segmentation performance by refining mask alignment but does not enhance morphological fidelity. Setting 3 employs Online-Augment to expand the training data, thereby enhancing both image quality and segmentation performance. Setting 4 integrates $\mathcal{L}_c$, boosting the performance of SANet \cite{SANet}, a texture-sensitive CNN model \cite{Intriguing}, and surpassing the previous SOTA model, ArSDM \cite{ArSDM}. Settings 5 through 7 demonstrate the synergistic effects of combining components. DHI acts as a catalyst for Online-Augment and $\mathcal{L}_c$, substantially improving image quality in Setting 6 and segmentation performance in Setting 7, thoroughly surpassing ArSDM. The final model, which integrates all components, achieves optimal results in both image fidelity and segmentation performance.

As discussed in \cref{Segmentation Comparison}, statistical quality and segmentation performance are not always directly correlated. Metrics like FID capture high-dimensional data distributions but tend to overfit with scarce data \cite{Anomalydiffusion} and miss fine-grained morphological details \cite{CMMD}, crucial for segmentation. Consequently, while $\mathcal{L}_c$ significantly improves segmentation performance by refining morphological characteristics, such as surface textures and color representation in Settings 4 and 7, its quality assessment results are lower than those of Online-Augment in Settings 3 and 6.

\textbf{Qualitative Analysis:} \cref{fig6} depicts the impact of each component. Settings 2 through 4 provided marginal improvements when applied individually. Setting 2 improves lesion-mask alignment, as shown in the first two columns. Setting 3 enhances color representation, while Setting 4 refines morphological texture features. In contrast, Settings 5 through 7, which combine components, significantly enhance morphological details. Our \textsl{Siamese-Diffusion}, integrating all components, generates the most realistic images with intricate details, demonstrating its efficacy.

\begin{table}[!t]\small
    \centering
    \caption{Comparison of the impact of different values of $w_{c}$ on medical image segmentation performance, evaluated using weighted mDice (\%) and mIoU (\%) metrics across three datasets. The training dataset consists of real images and images synthesized by our proposed \textsl{Siamese-Diffusion} method for each $w_c$.}\label{tab6}
    \setlength{\tabcolsep}{0.5mm}
    \scalebox{0.6}
    {\begin{tabular}{l|c c|c c|c c|c c|c c|c c|c c}
        \toprule[2pt]
        \multirow{4}{*}{$w_{c}$} & \multicolumn{6}{c|}{Polyps} & \multicolumn{4}{c|}{ISIC2016} & \multicolumn{4}{c}{ISIC2018} \\ 
        \cmidrule{2-15}
        ~ & \multicolumn{2}{c|}{SANet} & \multicolumn{2}{c|}{Polyp-PVT} & \multicolumn{2}{c|}{CTNet} & \multicolumn{2}{c|}{UNet} & \multicolumn{2}{c|}{SegFormer} & \multicolumn{2}{c|}{UNet} & \multicolumn{2}{c}{SegFormer} \\ 
        \cmidrule{2-15}
        ~ & mDice & mIoU & mDice & mIoU & mDice & mIoU & mDice & mIoU & mDice & mIoU & mDice & mIoU & mDice & mIoU \\
        \midrule[0.75pt]
        0.0 & 80.5 & 72.8 & 82.5 & 75.1 & 83.7 & 76.6 & 89.42 & 86.51 & 93.54 & 88.83 & 81.76 & 77.94 & 91.52 & 86.19 \\
        0.5 & 82.0 & 74.4 & \underline{84.0} & \underline{76.7} & 84.2 & 77.0 & 89.88 & 86.98 & 94.02 & 89.72 & 82.24 & 78.52 & 91.75 & 86.65 \\
        1.0 & \textbf{83.0} & \textbf{75.8} & \textbf{84.4} & \textbf{77.3} & \textbf{85.1} &\textbf{78.1} & \underline{89.91} & \underline{87.01} & \textbf{94.14} & \textbf{89.76} & \textbf{82.81} & \textbf{79.04} & \underline{91.80} & \underline{86.67} \\
        1.5 & \underline{82.4} & \underline{75.1} & \underline{84.0} & \underline{76.7} & \underline{84.6} & \underline{77.7} & \textbf{89.96} & \textbf{87.16} & 94.07 & 89.71 & \underline{82.41} & \underline{78.56} & \textbf{92.05} & \textbf{87.19} \\
        2.0 & 81.4 & 73.7 & 83.8 & 76.4 & 83.8 & 76.5 & 89.77 & 86.82 & \underline{94.11} & \underline{89.74} & 82.27 & 78.46 & 91.54 & 86.30 \\
        \bottomrule[2pt]
    \end{tabular}}
\end{table}

\subsubsection{Weight Adjustment of Noise Consistency Loss}\label{5.3.2}
Similar to CFG \cite{Classifier-free}, our method is akin to Image-Free Guidance (IFG) in the parameter space, with $w_c$ serving as the guidance strength. In this section, we investigate the impact of $w_c$ for $\mathcal{L}_c$ on the performance of various segmentation models, as shown in \cref{tab6}. We evaluate weights of $0.0$, $0.5$, $1.0$, $1.5$, and $2.0$. Setting $w_c = 0.0$ corresponds to ControlNet \cite{ControlNet}. When $w_c$ is set to $0.5$, $1.0$, or $1.5$, the mDice and mIoU scores of the segmentation models are consistently on par with or exceed those of the SOTA models. These results validate the effectiveness of the proposed \textsl{Noise Consistency Loss} $\mathcal{L}_c$ in enhancing segmentation performance and demonstrate the robustness of our approach. Finally, we select $w_c = 1.0$ as the optimal stable default.

\begin{figure}[!t]
    \centering
    \centerline{\includegraphics[width=\columnwidth]{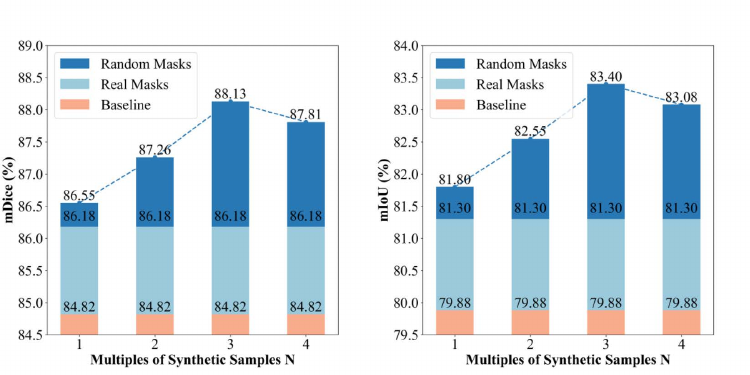}}
    \caption{Visualizing the trends of mDice (\%) and mIoU (\%) metrics for SegFormer \cite{SegFormer}. The training dataset consists of real polyp images and synthetic polyp images generated at different multiples by our proposed \textsl{Siamese-Diffusion}.}
    \label{fig7}
\end{figure}

\subsection{Discussion}
While the advantages of the mask-only sampling paradigm in terms of morphological diversity and scalability are preliminarily demonstrated in \cref{tab3} and \cref{tab4}, we further investigate the impact of varying data volumes on segmentation performance using SegFormer \cite{SegFormer}. As shown in \cref{fig7}, when the synthetic data volume equals one fold, images generated with transformed masks (\eg, scaling) show more pronounced improvements compared to those generated with real masks. This advantage continues to escalate as the data volume increases, peaking at threefold, where the mDice and mIoU metrics reach impressive values of $88.13\%$ and $83.40\%$. These results demonstrate that both scalability and morphological diversity are critical factors for segmentation. However, when the augmented data volume increases to fourfold, a decline in performance gains is observed. This intriguing phenomenon may be attributed to a unique ``long-tail problem" arising from the complex distribution within the Polyps training and testing datasets, which we plan to explore further in future work.

\section{Conclusion}\label{sec:conclusion}
In this paper, we introduce a novel image synthesis method, \textsl{Siamese-Diffusion}, which enhances the morphological fidelity of medical images while preserving the diversity inherent in the mask-only sampling paradigm. By employing \textsl{Noise Consistency Loss}, we generate high-fidelity medical images rich in morphological characteristics using arbitrary masks during the sampling phase. Extensive experiments across five datasets demonstrate the effectiveness and superiority of our method.
\section*{Acknowledgement}\label{sec:acknowledgement}
This work was supported in part by the Startup Grant for Professors (SGP) — CityU SGP, City University of Hong Kong, under Grant 9380170; partially by the Young Scientists Fund of the National Natural Science Foundation of China (Grant No. 62302328) and the Jiangsu Province Foundation for Young Scientists (Grant No. BK20230482); and partially by the WKU 2025 International Collaborative Research Program (Grant No. ICRPSP2025001). 
{
    \small
    \bibliographystyle{ieeenat_fullname}
    \bibliography{ref}
}

\clearpage
\setcounter{page}{1}
\maketitlesupplementary

\section{Additional Explanation of Architecture}
In this section, we provide additional preliminary knowledge explanations for \cref{sec:method} and present a detailed training algorithm of \textsl{Siamese-Diffusion}.
\label{sec:rationale}
\subsection{Architecture of Siamese-Diffusion}
Numerous methods \cite{ControlNet, T2I, Prompt-free} based on pre-trained Stable Diffusion \cite{SD} achieve fine control over synthesized images using external HyperNetworks \cite{Hypernetworks} to process structural inputs like segmentation masks, depth maps, and sketches. T2I-Adapter \cite{T2I} enables control with a lightweight feature extractor without updating Stable Diffusion’s parameters. Meanwhile, ControlNet \cite{ControlNet} demonstrates that precise control over synthesized images is achievable regardless of whether the denoising U-Net decoder parameters are updated. The key to the success of these methods lies in ensuring the effective fusion of prior control features and noisy image features in the latent space.

Although ControlNet \cite{ControlNet} demonstrates that updating the denoising U-Net decoder parameters is not essential for achieving fine control, such updates can improve the quality of synthesized images. In our proposed \textsl{Siamese-Diffusion} framework, \textsl{Mask-Diffusion} leverages collaborative updates from both the external network (comprising the cascaded Dense Hint Input module and ControlNet) and the denoising U-Net decoder parameters. This collaboration ensures the effective fusion of mask features $c_m$ with noisy image features $z_t$ in the latent space, enabling \textsl{Mask-Diffusion} to operate independently during sampling.

\textsl{Image-Diffusion} relies exclusively on the external feature extraction network for unidirectional fusion with noisy image features in the latent space. By leveraging \textsl{Noise Consistency Loss}, the U-Net decoder parameters are adjusted, enhancing the fusion between mask-image joint prior control features $c_{mix}$ and noisy image features $z_t$. As a result, \textsl{Image-Diffusion}, benefiting from the added image prior control, achieves more accurate noise predictions compared to \textsl{Mask-Diffusion}. \textsl{Noise Consistency Loss} further propagates these benefits to \textsl{Mask-Diffusion}, refining its parameters and enabling it to independently synthesize images with enhanced morphological characteristics. The ``copy” operation in \cref{fig2}(a) and the shared $c_m$ in \cref{eq5} ensure that the differences in noise predictions between the two processes are solely due to the additional image prior control, stabilizing the propagation process.

\begin{algorithm}[!t]
\caption{Training algorithm of Siamese-Diffusion.}\label{alg1}
\begin{algorithmic}[1]
    \For {Every batch of size N}
        \For {$(x_0, y_0)$ in this batch}
            \State Sample: \\
            \quad\quad\quad $\epsilon \sim \mathcal{N}(\mathbf{0}, \mathbf{I})$, $t\sim \mathcal{U}(\{0, 1, ..., T\})$
            \State Encode Image into Latent Space: \\
            \quad\quad\quad $z_{0}=\mathcal{E}(x_0)$
            \State Encode Prior Controls into Latent Space: \\
            \quad\quad\quad $c_{i}=\mathcal{F}(x_0)$, $c_{m}=\mathcal{F}(y_0)$ 
            \State Mix Mask and Image Prior Controls: \\
            \quad\quad\quad $c_{mix}=w_i \cdot c_{i}+w_m \cdot \textsl{sg}[c_{m}]$
            \State Noise Image in the Latent Space: \\
            \quad\quad\quad $z_{t} = \sqrt{\bar{\alpha}_{t}} z_{0}+ \sqrt{1-\bar{\alpha}_{t}}\epsilon$
            \State Calculate Mask Denoising Loss: \\ 
            \quad\quad\quad $\mathcal{L}_{m} = \| \epsilon_{\theta}(z_{t}, c_m) - \epsilon\|^2$
            \State Copy the Parameter of Mask-Diffusion: \\
            \quad\quad\quad $\theta^{\prime}=\text{DeepCopy}(\theta)$
            \State Calculate Image Denoising Loss: \\ 
            \quad\quad\quad $\mathcal{L}_{i} = \| \epsilon_{\theta^{\prime}}(z_{t}, c_{mix}) - \epsilon\|^2$
            \State Calculate Noise Consistency Loss: \\ 
            \quad\quad\quad $\mathcal{L}_{c} = w_c \cdot \|\epsilon_{\theta}(z_{t}, c_{m}) - \textsl{sg}[\epsilon_{\theta^{\prime}}(z_{t}, c_{mix})]\|^2$
            \State Single-Step Sampling: \\
            \quad\quad\quad $z_0^{\prime} = \frac{z_t - \sqrt{1 -\Bar{\alpha}_t}\textsl{sg}[\epsilon_{\theta^{\prime}}(z_{t}, c_{mix})]}{\sqrt{\Bar{\alpha}_t}}$
            \State Noise Image in the Latent Space: \\
            \quad\quad\quad $z_{t}^{\prime} = \sqrt{\bar{\alpha}_{t}} z_{0}^{\prime}+ \sqrt{1-\bar{\alpha}_{t}}\epsilon$
            \State Calculate Augmented Mask Denoising Loss: \\
            \quad\quad\quad $\mathcal{L}_{m^{\prime}} = w_a \cdot \| \epsilon_{\theta}(z_{t}^{\prime}, c_{m}) - \epsilon\|^2$
            \State Update the Parameters: \\
            \quad\quad\quad $\mathcal{L} = \mathcal{L}_m +\mathcal{L}_i + \mathcal{L}_c + \mathcal{L}_{m^{\prime}}$
        \EndFor
    \EndFor
\end{algorithmic}
\end{algorithm}

\textsl{Mask-Diffusion} and \textsl{Image-Diffusion} share a unified diffusion model. The Siamese architecture stands out from traditional distillation models \cite{Distilling, DistillationClassifier}, which require separate training of a teacher network with mask-image prior control and a student network with mask-only prior control. Thus, the Siamese structure relies on fewer parameters, making it more resource efficient for training.

\subsection{Dense Hint Input Module}
As mentioned above, simultaneously updating the denoising U-Net parameters enhances the quality of synthesized images. However, \textsl{Image-Diffusion} compromises the fusion between image prior control features $c_i$ and noisy image features $z_t$. In \cite{Prompt-free}, high-density semantic image features achieve unidirectional fusion with noisy image features using a robust external network, without requiring collaborative updates to the denoising U-Net parameters. To enhance the fusion between mask-image joint prior control features $c_{mix}$ and noisy image features $z_t$, we substitute the original sparse Hint Input module with a more powerful Dense Hint Input module. Furthermore, \cite{Prompt-free} demonstrates that a stronger external HyperNetwork can effectively accommodate low-density semantic inputs (\eg, segmentation masks). As a result, \textsl{Mask-Diffusion} and \textsl{Image-Diffusion} share a unified external feature extractor in our approach.

\subsection{Online-Augmentation}
Typically, due to the low accuracy of noise prediction in single-step sampling, diffusion models (\eg, \cite{DDPM,DDIM,Score-based}) rely on Predict-Correct (PC) multi-step sampling. However, using multi-step sampling during training is both time-consuming and memory-intensive, as it requires storing gradient information \cite{ControlNetPlus}, which limits the feasibility of generating new sample pairs online to augment the training set. By leveraging the advantages of the \textsl{Siamese-Diffusion} training paradigm, single-step sampling can be used for online augmentation in \textsl{Mask-Diffusion}.

\subsection{Training Algorithm of Siamese-Diffusion}
The detailed training algorithm of \textsl{Siamese-Diffusion} is presented in \cref{alg1}. For simplicity, the prompt $c_t$ and timestep $t$ are omitted. $\mathcal{E}$ represents the encoder of VQ-VAE \cite{VQVAE}, whose parameters are frozen. $\mathcal{F}$ represents the external feature extraction network, consisting of the cascaded Dense Hint Input module and ControlNet, which are jointly trained with the Diffusion model.

\section{Test Set for Segmentation Models} This section provides detailed descriptions of the test sets used to evaluate segmentation models across various datasets, as discussed in \cref{sec:dataset}.

\textbf{Polyps Dataset:} Following \cite{ArSDM}, evaluations are conducted on five public polyp datasets: EndoScene \cite{EndoScene} (60 samples), CVC-ClinicDB/CVC-612 \cite{CVC-ClinicDB} (62 samples), Kvasir \cite{Kvasir} (100 samples), CVC-ColonDB \cite{CVC-ColonDB} (380 samples), and ETIS \cite{ETIS} (196 samples). \textsl{Overall} represents the weighted average results of these five test sets.

\textbf{ISIC2016 \& ISIC2018 Datasets:} Evaluations are performed on their official test sets.

\textbf{Stain \& Feces Datasets:} As described in the main paper, the Stain dataset (500 samples) and the Faeces dataset (458 samples) are divided into training, validation and test sets in a 3:1:1 ratio, resulting in test sets of 100 and 92 samples.

\begin{table}[!t]\small
    \centering
    \caption{Comparison of synthetic skin lesion (ISIC2016) image quality generated by each respective mask-only method, evaluated using FID \cite{FID}, KID \cite{KID}, CLIP-I \cite{CLIP-I}, LPIPS \cite{LPIPS}, CMMD \cite{CMMD} and MOS metrics.}
    \label{tab7}
    \setlength{\tabcolsep}{1.0mm}
    \scalebox{0.7}
    {\begin{tabular}{l|c c c c c c}
        \toprule[2pt]
        Methods & FID ($\downarrow$) & KID ($\downarrow$) & CLIP-I ($\uparrow$) & LPIPS ($\downarrow$) & CMMD ($\downarrow$) & MOS ($\uparrow$) \\ 
        \midrule[0.75pt]
        T2I-Adapter \cite{T2I} & 234.474 & 0.1912 & 0.774 & 0.688 & 2.733 & - \\
        ControlNet \cite{ControlNet} & 68.327 & \textbf{0.0267} & 0.820 & 0.667 & \textbf{0.688} & 1.68\\
        \textbf{Ours} & \textbf{64.208} & 0.0299 & \textbf{0.827} & \textbf{0.657} & 0.733 & \textbf{1.96}\\
        \bottomrule[2pt]
    \end{tabular}}
\end{table}

\begin{table}[!t]\small
    \centering
    \caption{Comparison of synthetic skin lesion (ISIC2018) image quality generated by each respective mask-only method, evaluated using FID \cite{FID}, KID \cite{KID}, CLIP-I \cite{CLIP-I}, LPIPS \cite{LPIPS}, CMMD \cite{CMMD} and MOS metrics.}
    \label{tab8}
    \setlength{\tabcolsep}{1.0mm}
    \scalebox{0.7}
    {\begin{tabular}{l|c c c c c c}
        \toprule[2pt]
        Methods & FID ($\downarrow$) & KID ($\downarrow$) & CLIP-I ($\uparrow$) & LPIPS ($\downarrow$) & CMMD ($\downarrow$) & MOS ($\uparrow$) \\ 
        \midrule[0.75pt]
        T2I-Adapter \cite{T2I} & 224.446 & 0.1751 & 0.796 & \textbf{0.672} & 2.621 & - \\
        ControlNet \cite{ControlNet} & 45.490 & 0.0286 & \textbf{0.809} & \textbf{0.672} & 0.794 & 1.22 \\
        \textbf{Ours} & \textbf{44.036} & \textbf{0.0258} & 0.808 & 0.673 & \textbf{0.701} & \textbf{2.12}\\
        \bottomrule[2pt]
    \end{tabular}}
\end{table}

\section{Image Quality Comparison}
This section provides additional quantitative and qualitative analysis of image quality assessment results on other datasets, as discussed in \cref{sec:imagequality}.
\subsection{Quantitative Evaluation}
For the five datasets, we use various image quality evaluation metrics, including Fréchet Inception Distance (FID) \cite{FID}, Kernel Inception Distance (KID) \cite{KID}, CLIP-Image (CLIP-I) \cite{CLIP-I}, Learned Perceptual Image Patch Similarity (LPIPS) \cite{LPIPS} and CLIP-Maximum Mean Discrepancy (CMMD) \cite{CMMD}. Additionally, for medical datasets, we employ the Mean Opinion Score (MOS), calculated by averaging experienced clinicians’ ratings of synthetic image quality. Distinct evaluation standards are applied to the Polyps and Skin Lesion datasets based on clinician suggestions

\textbf{Polyps Dataset:} As shown in \cref{tab1}, the commonly used non-human evaluation metrics have been discussed in the main paper. Here, we focus on the MOS metric. Three professional clinicians assess the quality of polyp images generated by SinGAN-Seg \cite{Singan}, ControlNet \cite{ControlNet}, our \textsl{Siamese-Diffusion}, and real data. T2I-Adapter \cite{T2I} and ArSDM \cite{ArSDM} are excluded based on clinicians’ suggestions due to their unrealistic results. To ensure the reliability of the evaluation results and minimize the fatigue of the clinicians, $50$ image groups are randomly selected, each containing images synthesized using the same mask by the four methods (including real images), totaling $200$ polyp images. Clinicians view one image at a time, scoring it as ``Real" ($1$ point) or ``Synthetic" ($0$ point) and providing a confidence score ranging from $1$ to $10$. Our method achieves the highest MOS score of $0.587$ with a confidence level of $6.04$, demonstrating superior quality in synthesized polyp images. $50$ real images are used to assess preference, yielding MOS score of $0.9$ and confidence level of $6.17$.

\textbf{ISIC2016 Dataset:} As shown in \cref{tab7}, we compare the synthesized skin lesion image quality of T2I-Adapter \cite{T2I}, ControlNet \cite{ControlNet}, and our \textsl{Siamese-Diffusion}. Although KID \cite{KID} and CMMD \cite{CMMD} scores are slightly lower than those of ControlNet \cite{ControlNet}, our method achieves the best scores in FID \cite{FID}, CLIP-I \cite{CLIP-I}, LPIPS \cite{LPIPS}, and MOS, demonstrating superior overall performance. About MOS evaluation, one experienced clinician assesses the quality of skin lesion images generated by ControlNet \cite{ControlNet}, our \textsl{Siamese-Diffusion}, and real data. T2I-Adapter \cite{T2I} is excluded for the same reason as in the Polyps dataset. Considering potential biases in single-evaluator assessments, these results are presented as reference points rather than definitive benchmarks. Following clinician suggestions, $50$ image groups are randomly selected, each containing images synthesized using the same mask by the three methods (including real images), totaling $150$ skin lesion images. The clinician views three images at a time and assigns rankings from $1$ to $3$ (highest). Our method achieves the highest MOS score of $1.96$. $50$ real images are used to assess preference, yielding an MOS score of $2.36$.

\textbf{ISIC2018 Dataset:} As shown in \cref{tab8}, similar to the ISIC2016 dataset, our method performs better overall. The ISIC2018 dataset ($2,594$ samples) contains significantly more trainable data compared to the ISIC2016 dataset ($900$ samples), resulting in markedly improved FID \cite{FID} and KID \cite{KID} scores. This observation supports the notion that increasing the amount of training data for generative models can improve the quality of synthesized images. Incredibly, T2I-Adapter \cite{T2I} and ControlNet \cite{ControlNet} achieve identical LPIPS \cite{LPIPS} scores of $0.672$, which are marginally better than the $0.673$ achieved by our \textsl{Siamese-Diffusion}. However, the visualization results in \cref{fig10}(b) contradict this outcome, suggesting that LPIPS \cite{LPIPS} may not reliably reflect alignment with human perception. The MOS evaluation standards applied to the ISIC2018 dataset are the same as those used for ISIC2016. Our method achieves the highest MOS score of $2.12$. Additionally, $50$ real images are used to assess preference, yielding an MOS score of $2.66$.

\textbf{Stain \& Faeces Datasets:} As shown in \cref{tab9} and \cref{tab10}, we compare the synthesized image quality of DFMGAN \cite{DFMGAN}, AnomalyDiffusion \cite{Anomalydiffusion}, T2I-Adapter \cite{T2I}, ControlNet \cite{ControlNet}, and our \textsl{Siamese-Diffusion}. Our method outperforms the others on both datasets, demonstrating the superiority of our approach. Shockingly, T2I-Adapter \cite{T2I} achieves the best CMMD \cite{CMMD} score on the Faeces dataset, but discrepancies with the visualization in \cref{fig11}(c) highlight potential limitations of the CMMD metric.

\begin{table}[!t]\small
    \centering
    \caption{Comparison of synthetic stain image quality generated by each respective mask-only method, evaluated using FID \cite{FID}, KID \cite{KID}, CLIP-I \cite{CLIP-I}, LPIPS \cite{LPIPS}, and CMMD \cite{CMMD} metrics.}
    \label{tab9}
    \setlength{\tabcolsep}{1.0mm}
    \scalebox{0.7}
    {\begin{tabular}{l|c c c c c}
        \toprule[2pt]
        Methods & FID ($\downarrow$) & KID ($\downarrow$) & CLIP-I ($\uparrow$) & LPIPS ($\downarrow$) & CMMD ($\downarrow$)\\ 
        \midrule[0.75pt]
        DFMGAN \cite{Singan} & 242.780 & 0.1619 & 0.712 & 0.781 & 3.733 \\
        AnomalyDiffusion \cite{ControlNet} & 165.732 & 0.0791 & 0.763 & 0.778 & 1.296 \\
        T2I-Adapter \cite{T2I} & 209.260 & 0.1371 & 0.765 & 0.778 & 1.296 \\
        ControlNet \cite{ArSDM} & 123.818 & 0.0298 & 0.769  & 0.731 & 1.213 \\
        \textbf{Ours} & \textbf{115.546} & \textbf{0.0206} & \textbf{0.773} & \textbf{0.719} & \textbf{1.183}\\
        \bottomrule[2pt]
    \end{tabular}}
\end{table}

\begin{table}[!t]\small
    \centering
    \caption{Comparison of synthetic faece image quality generated by each respective mask-only method, evaluated using FID \cite{FID}, KID \cite{KID}, CLIP-I \cite{CLIP-I}, LPIPS \cite{LPIPS}, and CMMD \cite{CMMD} metrics.}
    \label{tab10}
    \setlength{\tabcolsep}{1.0mm}
    \scalebox{0.7}
    {\begin{tabular}{l|c c c c c}
        \toprule[2pt]
        Methods & FID ($\downarrow$) & KID ($\downarrow$) & CLIP-I ($\uparrow$) & LPIPS ($\downarrow$) & CMMD ($\downarrow$)\\ 
        \midrule[0.75pt]
        DFMGAN \cite{Singan} & 299.032 & 0.2156 & 0.639 & 0.760 & 6.369 \\
        AnomalyDiffusion \cite{ControlNet} & 220.003 & 0.1181 & 0.733 & 0.754 & 2.232\\
        T2I-Adapter \cite{T2I} & 207.814 & 0.1118 & 0.778 & 0.651 & \textbf{1.264} \\
        ControlNet \cite{ArSDM} & 166.567 & 0.0843 & 0.765  & 0.651 & 1.701 \\
        \textbf{Ours} & \textbf{143.736} & \textbf{0.0485} & \textbf{0.786} & \textbf{0.643} & 1.337 \\
        \bottomrule[2pt]
    \end{tabular}}
\end{table}

\subsection{Qualitative Evaluation}
In this section, we provide a qualitative analysis of synthesized images from four datasets generated by different methods. For sensitivity reasons, synthesized images for the Faeces dataset are not displayed.

\begin{figure*}[!t]
    \centering
    \centerline{\includegraphics[width=\textwidth]{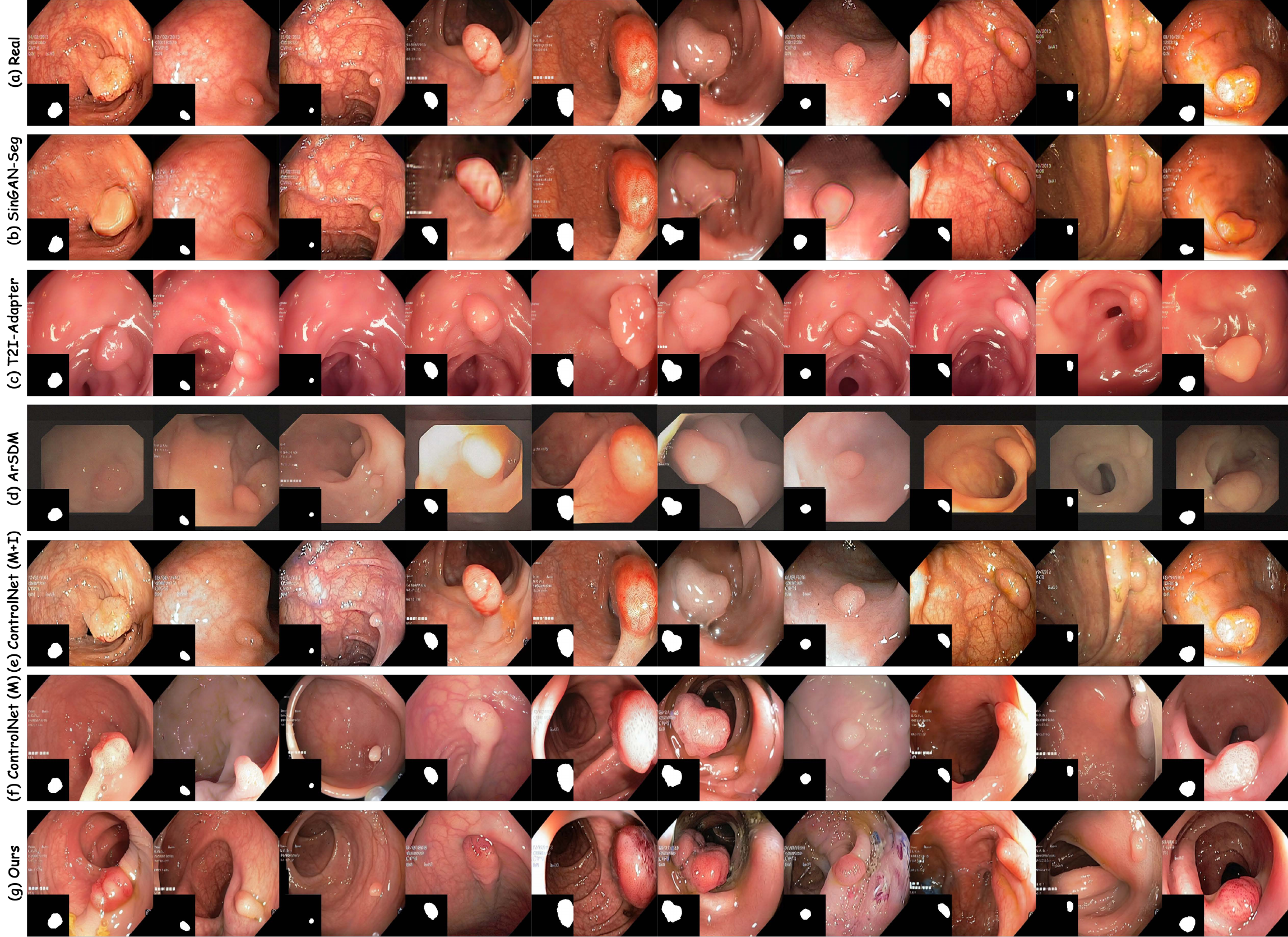}}
    \caption{(a) Examples of real polyp images. (b)–(g) Examples of synthetic polyp images generated by each respective method. ``M" denotes that mask-only prior control, while ``M+I" denotes mask-image joint prior control. The synthetic polyp images generated by our method achieve competitive fidelity while also exhibiting diversity (Zoom in for better visualization).}
    \label{fig8}
\end{figure*}

\textbf{Polyps Dataset:} 
\cref{fig8} presents visualizations of polyp images synthesized by different generative models. The differences between each method have been discussed in the main paper. Additional examples further demonstrate that our method produces images with rich morphological characteristics, validating the superiority of our approach. Notably, the ``editing-like" approach of SinGAN-Seg \cite{Singan} generates minimal artifacts when the mask varies slightly, aligning with human perception when viewed without zooming in. However, when the mask undergoes significant variations, the artifacts become extremely pronounced, undermining the realism of the synthesized images.

\begin{figure*}[!t]
    \centering
    \centerline{\includegraphics[width=\textwidth]{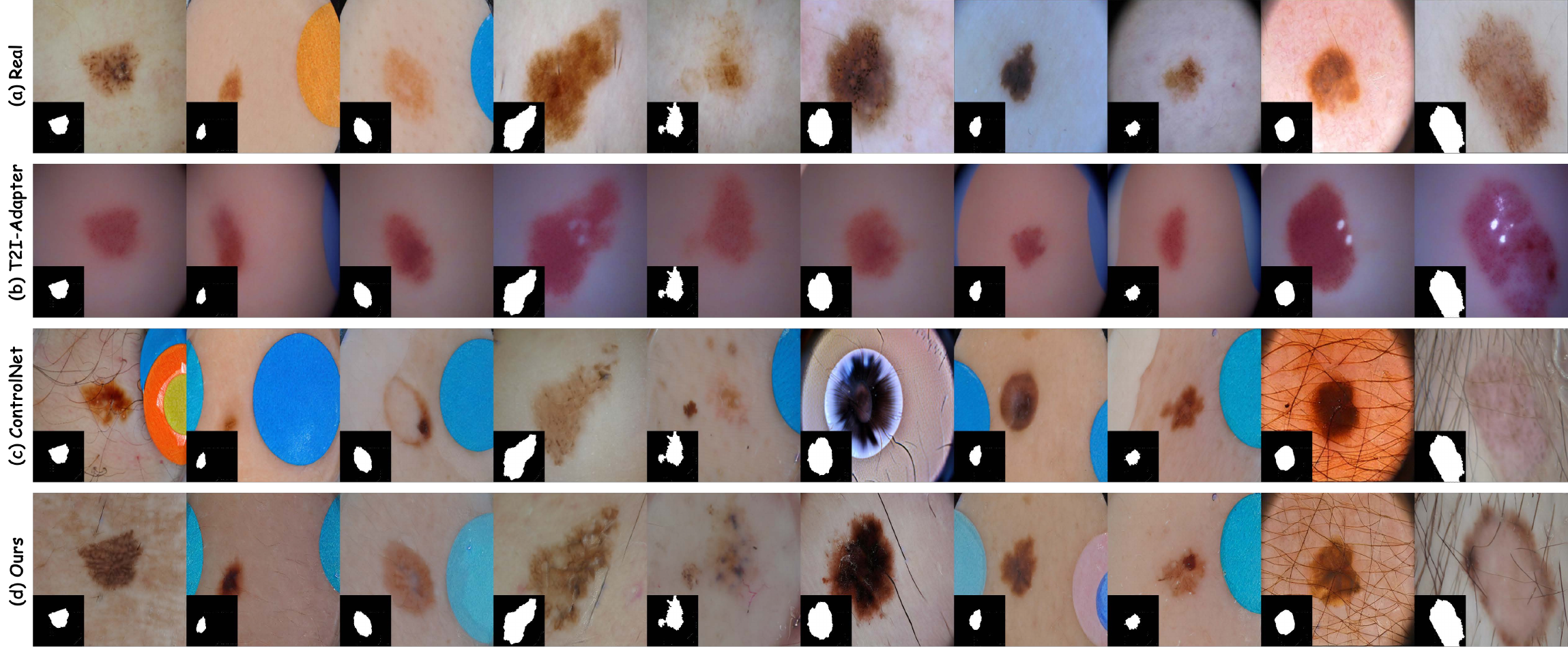}}
    \caption{(a) Examples of real skin lesion (ISIC2016) images. (b)–(d) Examples of synthetic skin lesion images generated by each respective method. The synthetic skin lesion images generated by our method achieve competitive fidelity while also exhibiting diversity (Zoom in for better visualization).}
    \label{fig9}
\end{figure*}

\begin{figure*}[!t]
    \centering
    \centerline{\includegraphics[width=\textwidth]{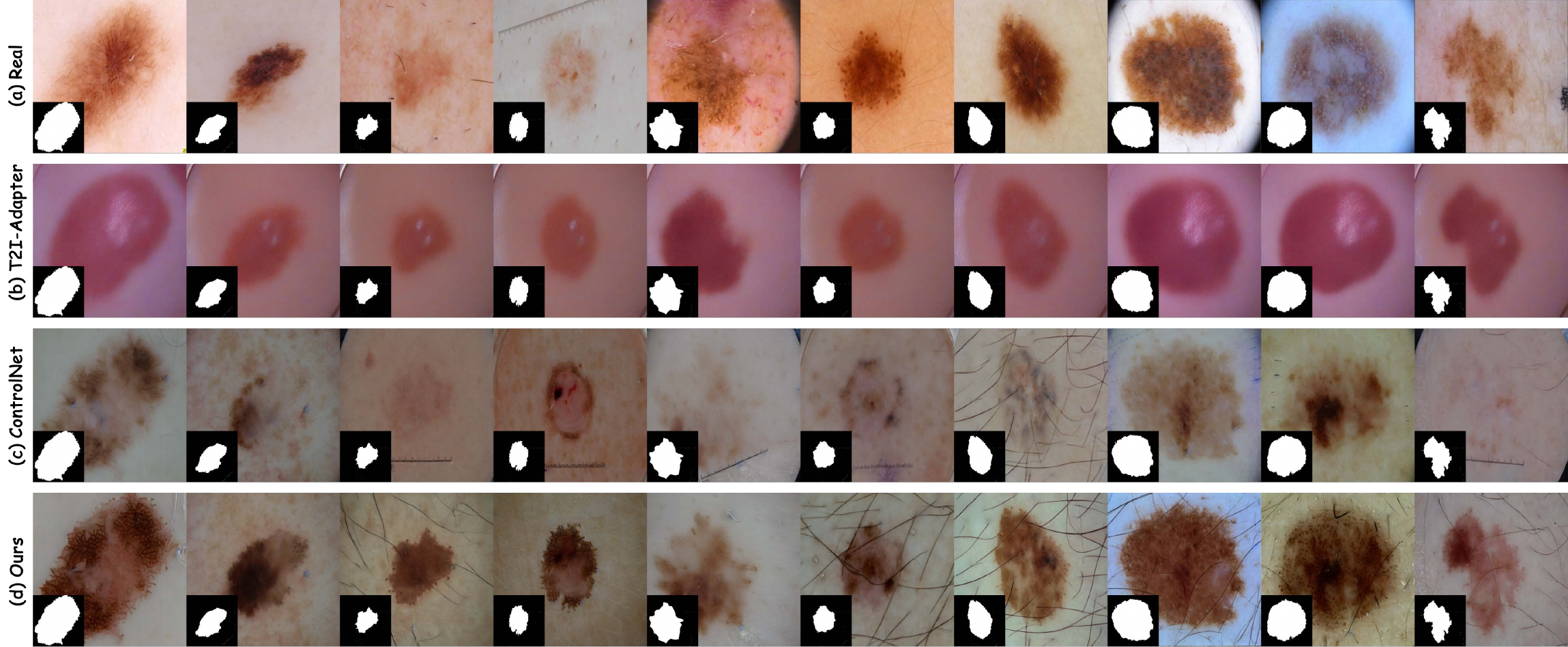}}
    \caption{(a) Examples of real skin lesion (ISIC2018) images. (b)–(d) Examples of synthetic skin lesion images generated by each respective method. The synthetic skin lesion images generated by our method achieve competitive fidelity while also exhibiting diversity (Zoom in for better visualization).}
    \label{fig10}
\end{figure*}

\textbf{ISIC2016 \& ISIC2018 Datasets:} \cref{fig9} and \cref{fig10} present visualizations of skin lesion images generated by various models, revealing phenomena similar to those observed in the Polyps dataset. T2I-Adapter \cite{T2I} generates unrealistic images with a uniform ``style". Compared to ControlNet \cite{ControlNet}, our method demonstrates superior performance in mask alignment, morphological texture, and color, validating the effectiveness and superiority of our approach.

\begin{figure*}[!t]
    \centering
    \centerline{\includegraphics[width=\textwidth]{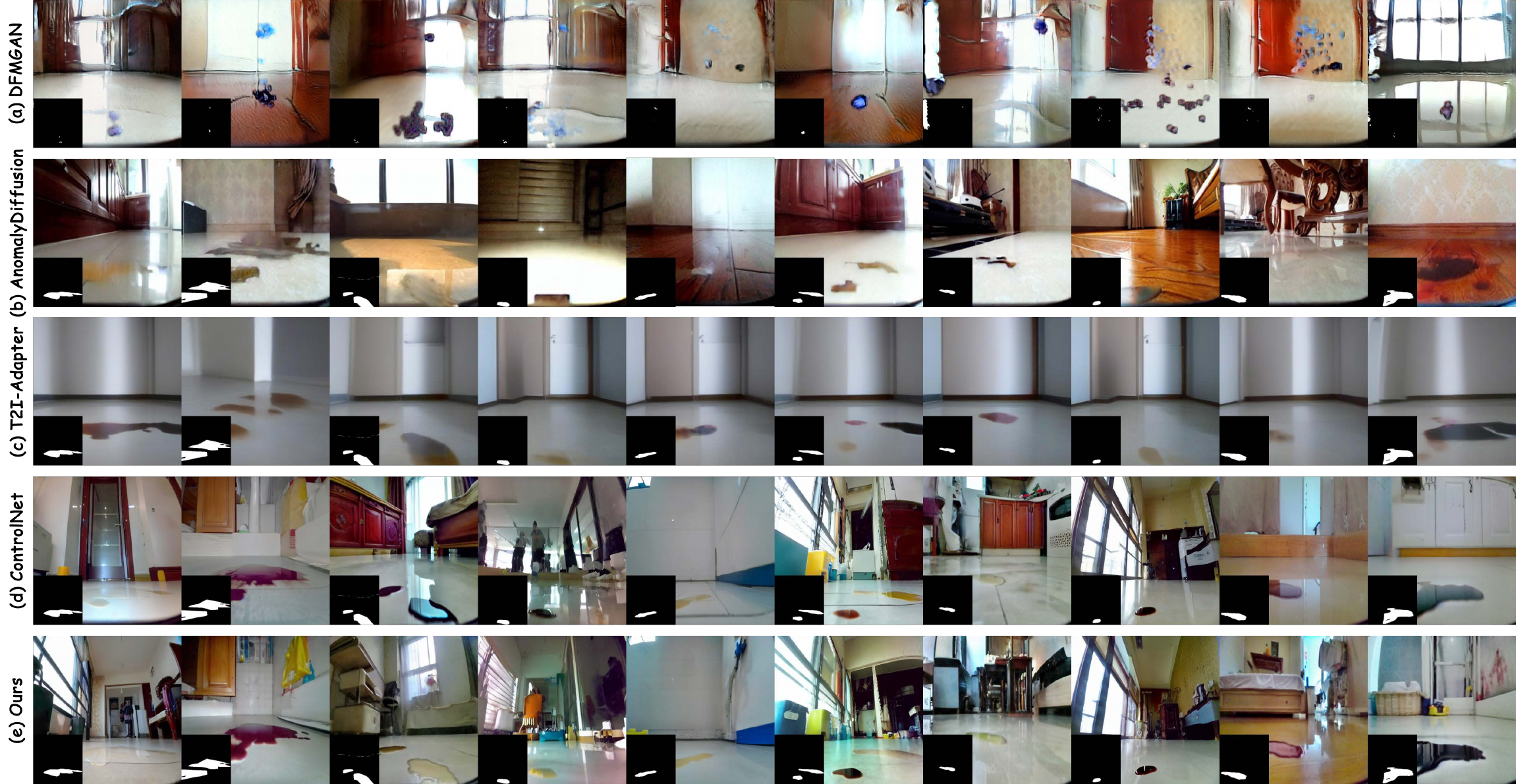}}
    \caption{(a)–(e) Examples of synthetic stain images generated by each respective method are shown. DFMGAN \cite{DFMGAN} cannot control the synthesis with the specified masks. The synthetic stain images generated by our method exhibit higher semantic density, meaning richer content, which corresponds to the rich morphological characteristics in medical images. Additionally, these synthetic stain images achieve competitive fidelity while also demonstrating diversity (Zoom in for better visualization).}
    \label{fig11}
\end{figure*}

\textbf{Stain Dataset:} \cref{fig11} presents visualizations of stain images synthesized by various models. DFMGAN \cite{DFMGAN} cannot control the synthesis with the specified masks and performs poorly when data is scarce. AnomalyDiffusion \cite{Anomalydiffusion} demonstrates poor alignment of the mask and stain, especially when the mask area is small. T2I-Adapter \cite{T2I} generates images with a unified background ``style" and exhibits low content density. In terms of content density, our method outperforms ControlNet \cite{ControlNet}, generating richer content, which corresponds to the richness of morphological characteristics in medical images, thus validating the effectiveness and superiority of our approach.

\section{Qualitative Analysis of Each Component}
\begin{figure*}[!t]
    \centering
    \centerline{\includegraphics[width=\textwidth]{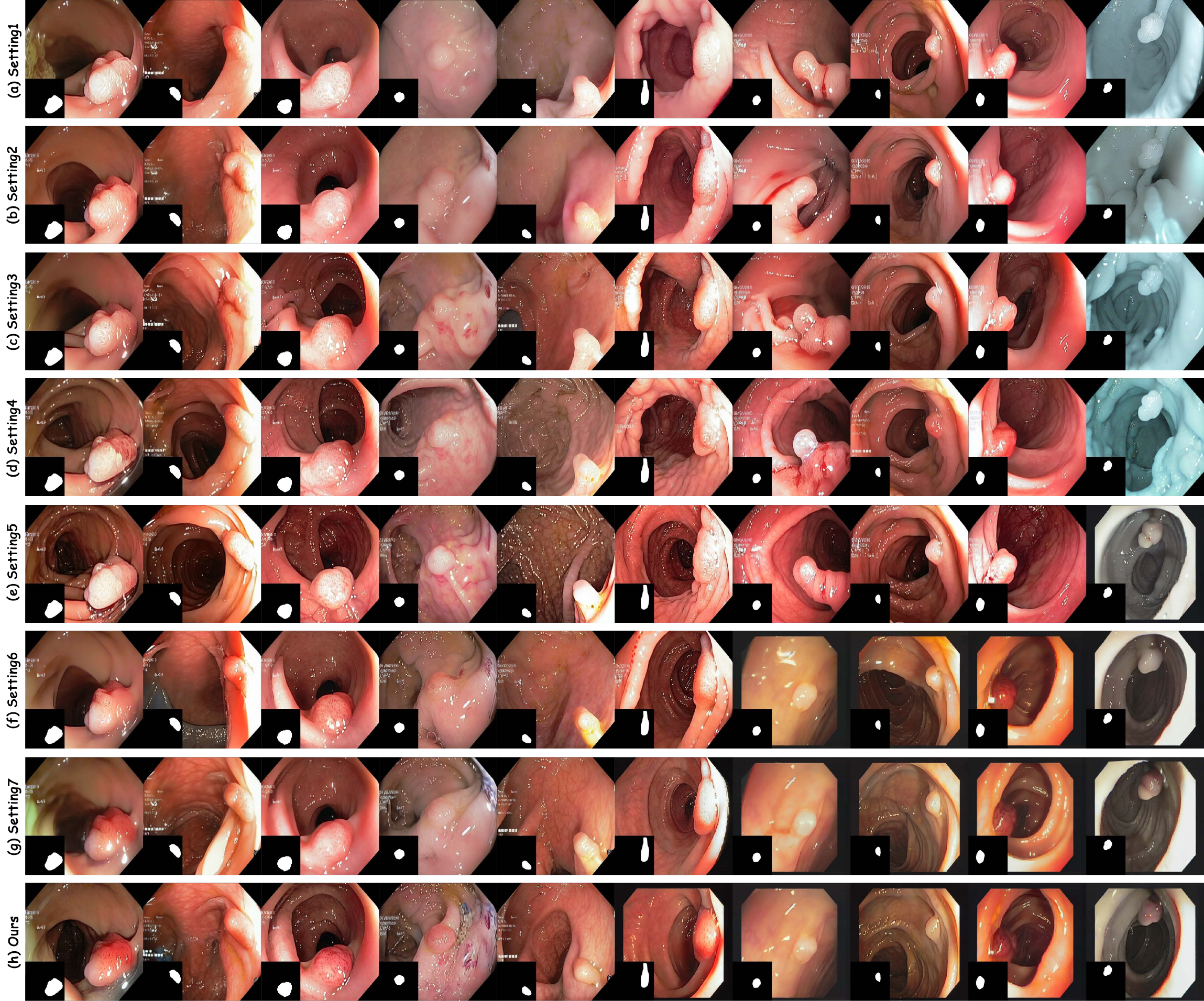}}
    \caption{Visualizing the impact of different components on the synthesis of polyp images (Zoom in for better visualization).}
    \label{fig12}
\end{figure*}

In this section, we present additional images illustrating the impact of each component, as shown in \cref{fig12}, to further substantiate the conclusions drawn in \cref{components}.

\end{document}